\documentclass[a4paper,fleqn]{cas-dc}

\usepackage{amsmath,amsfonts,amssymb}
\usepackage{algorithmic}
\usepackage{algorithm}
\usepackage{array}
\newcolumntype{P}[1]{>{\centering\arraybackslash}p{#1}}
\usepackage[caption=false,font=normalsize,labelfont=sf,textfont=sf]{subfig}
\usepackage{textcomp}
\usepackage{stfloats}
\usepackage{url}
\usepackage{verbatim}
\usepackage{graphicx}
\usepackage{cite}
\def\BibTeX{{\rm B\kern-.05em{\sc i\kern-.025em b}\kern-.08em
    T\kern-.1667em\lower.7ex\hbox{E}\kern-.125emX}}
\usepackage{booktabs}
\usepackage{multirow}
\usepackage{siunitx}
\usepackage[table]{xcolor}
\usepackage[table]{xcolor}
\usepackage{color,soul}
\sethlcolor{yellow}

\usepackage[numbers]{natbib}

\newtheorem{defka}{Definition}
\newtheorem{thm}{Theorem}

\newtheorem{lem}{Lemma}
\newtheorem{rmk}{Remark}
\newtheorem{prop}{Property}
\newtheorem{pf}{Proof}
 
\usepackage{mathrsfs}

\newcommand{\defequal}{\overset{\text{def}}{=}}
\begin{document}


\title [mode = title] { Robust Immersive Bilateral Teleoperation of Beyond-Human-Scale Manipulators with Enhanced Transparency and Sense of Embodiment}

\author[1]{Mahdi Hejrati} \cormark[1] 
\ead{mahdi.hejrati@tuni.fi} 

\author[1]{Pauli Mustalahti}
\ead{pauli.mustalahti@tuni.fi}

\author[1]{Jouni Mattila}
\ead{jouni.mattila@tuni.fi}

\affiliation[1]{organization={Department of Engineering and Natural Sciences, Tampere University}, city={Tampere}, country={Finland}} \cortext[1]{Corresponding author: Mahdi Hejrati}




\shorttitle{Robust Immersive Bilateral Teleoperation}
\shortauthors{Hejrati et al.}

\begin{abstract}
This paper presents an immersive bilateral teleoperation framework for beyond-human-scale manipulators, combining motion/force transparency with enhanced operator embodiment through virtual reality (VR) and distributed haptic feedback. The platform integrates a full-scale industrial hydraulic manipulator, a 7-DoF haptic exoskeleton, and head-tracked visual feedback to reinforce operator agency and self-location. A force-sensorless adaptive controller incorporates an augmented human–robot dynamic model to address master–surrogate asymmetries, input nonlinearities, model uncertainties, and communication delays. Rigorous analysis establishes semi-global uniform ultimate boundedness of the closed-loop system. Extensive full-scale experiments demonstrate precise motion and force tracking with scaling up to 1:13 and 1:1000, respectively, and fixed or time-varying delays up to 150 ms. Coordinated 6-DoF pick-and-place and multi-axis contact tasks further validate system performance. A ten-participant user study supports system usability and indicates approximately 50\% higher normalized embodiment scores with VR than with monitor-based feedback. These results advance intuitive, force-reflected remote manipulation at industrial scale.

\end{abstract}

\begin{keywords}
Bilateral teleoperation control\sep Adaptive control\sep Transparency\sep Virtual reality\sep Haptic display\sep Sense of embodiment
\end{keywords}

\maketitle

\section{Introduction}
Teleoperated robotic systems, first introduced by Goertz in the 1940s \cite{hokayem2006bilateral}, integrate human cognitive abilities with control-theoretic approaches to enable remote operation of robots in environments that are hazardous, unstructured, or otherwise unsuitable for direct human presence. In scenarios where use of fully autonomous systems remain impractical or unsafe, teleoperation offers a robust and flexible alternative, with proven effectiveness across a wide range of domains \cite{liu2023multi,chen2019mode, su2019improved, selvaggio2021shared}. One field with significant untapped potential for teleoperation is heavy-duty robotic operation, particularly through the use of heavy-duty hydraulic manipulators (HHMs) \cite{liu2023multi, le2017remote, shen2024hybrid}. These machines are already widely deployed in industries such as mining, construction, and forestry, where they handle physically intensive tasks in dynamic and often dangerous environments \cite{cho2026collaborative}. By integrating human sensorimotor skills with the power and scale of HHMs, teleoperation equips HHMs with beyond human capabilities, while providing a reliable medium for intuitive human-to-HHM skill transfer \cite{suomalainen2018learning}, addressing a key bottleneck in their automation \cite{mattila2017survey}.

Despite the advantages of teleoperation, task performance often deteriorates during remote operation due to factors such as master-surrogate mismatches, communication delays, and non-intuitive human-machine interfaces that can cause cyber sicknesses \cite{zhang2018detection}. These challenges become even more pronounced when teleoperating HHMs, whose scale, dynamics, and mechanical complexity make them significantly less tractable than lightweight robotic systems. Such issues disrupt the operator's natural sensorimotor loop and can impair proprioceptive perception, especially with visuo-haptic feedback, leading to a loss of intuitiveness in the teleoperation experience. One effective approach to mitigate these effects is enhancing the operator’s sense of \textit{presence} \cite{slater1997framework}, defined as the subjective experience of being physically situated at the remote site, which, in the context of teleoperation, is more precisely referred to as the sense of \textit{telepresence} (SoT) \cite{ellis1996presence, nash2000review}. Nonetheless, enhancing telepresence alone is not sufficient to ensure high-quality task execution in teleoperation by a human operator.

Slater (2009) \cite{slater2009place} argued that Place Illusion—his proposed refinement of the term "presence"—is not sufficient on its own for users to behave realistically in virtual environments. He introduced the concept of Plausibility Illusion, defined as the illusion that events occurring in the virtual environment are actually happening. According to Slater, both Place Illusion and Plausibility Illusion must co-occur for users to generate realistic responses within the virtual context. Building on this framework, Kilteni et al. (2012) \cite{kilteni2012sense} emphasized that realistic actions in virtual environments are closely linked to the sense of embodiment (SoE). Within the concept of teleoperation, SoE is defined as a psychological state in which the operator experiences the surrogate as an extension of their own body. According to \cite{kilteni2012sense,falcone2023toward}, SoE encompasses three critical components: (1) \textit{body ownership}—feeling the surrogate as part of oneself; (2) \textit{self-location}—a strong spatial telepresence at the remote site; and (3) \textit{agency}—the confidence that one's intentions directly produce corresponding actions in the remote environment. In many real-world applications—particularly industrial and field scenarios—the most relevant embodiment components are agency and self-location \cite{falcone2023toward}, which directly influence the operator's control quality and task performance. 

Thus, a strong SoE is crucial for successful teleoperation. When the SoE is strong, the operator no longer perceives the surrogate,the teleoperated robot, as an external tool, but rather as an extension of their own body and sensorimotor system \cite{cabrera2017human}. This shift minimizes cognitive load and enhances the fluidity of task execution. Immersion is one key to evoking this SoE. Through the combination of visual feedback, delivered via a virtual reality (VR) headset, and distributed haptic feedback from force-reflective exoskeletons, immersive bilateral teleoperation creates more realistic task operation. But for immersion to truly support embodiment, it must be grounded in two essential pillars, each addressing relevant components of SoE: precise control, which reinforces the operator’s sense of agency, and appropriate equipment, which anchors their sense of self-location. This study addresses both dimensions in the context of industrial surrogates, focusing on HHMs. By developing a robust, high-accuracy control framework and pairing it with immersive, well-integrated hardware, we aim to strengthen both agency and self-location. This enables HHMs to combine human-like precision with mechanical strength, empowering operators to perform complex tasks efficiently and safely in hazardous or unstructured environments.

\section{Related Works}
\subsection{Sense of Self-Location}
Immersive bilateral teleoperation relies on the integration of multi-modal sensory feedback, including visual, auditory, and haptic cues, to foster a stronger connection between the human operator and the remote environment. This immersive experience enhances the SoT, which in turn reinforces the sense of self-location (SoSL), as the two are closely interlinked. It is important to note, however, that not all sensory inputs contribute equally to immersion. Visual feedback, in particular, plays a foundational role in shaping the immersive experience during teleoperation \cite{wei2021multi}. Yet, immersion itself is a qualitative and often comparative concept \cite{slater2009place}; one system can be considered as more immersive than another with the given visual feedback, if, for instance, it provides real-time egocentric head-tracking compared to one with static camera. Even within the same system configuration, enhancements such as higher display resolution or the addition of haptic feedback can further elevate the sense of immersion, and, by extension, strengthen the operator’s SoSL.

Building upon this foundation, Cheng et al. \cite{cheng2024open} introduced Open-TeleVision, an immersive teleoperation system that allows operators to intuitively perceive the robot's environment in a stereoscopic manner. Central to their approach is a stereo RGB camera mounted on the humanoid robot’s head, equipped with a mechanism that dynamically tracks the operator's head movements. This configuration enables real-time, egocentric visual feedback streamed directly to a VR device, effectively strengthening the operator's sense of being present in the remote environment. Transitioning from single-view perception to a more comprehensive spatial awareness, \cite{girbes2020haptic} proposes a system for dual-arm bilateral teleoperation that further boosts the SoE. Their method combines multi-view visual feedback with haptic interaction, offering the operator five distinct visual sections—each from different camera perspectives—alongside real-time force graphs and parameter reconfiguration. This multi-faceted feedback setup allows the operator to maintain a heightened situational awareness, contributing to more precise and confident task execution. The immersive teleoperation framework proposed in \cite{li2024reality} increased the task execution performance by balancing the trade-off between data streaming costs and data visual quality in immersive VR. Other papers, moreover, leveraged the enhanced VR to achieve higher degrees of immersion in teleoperation \cite{de2021leveraging,stotko2019vr}. Expanding on the integration of sensory modalities, \cite{huang2024telepresence} presents a visual-haptic perception and reconstruction system designed to enhance telepresence. Their system not only delivers visual information from the remote site through VR but also provides haptic guidance to the operator, fostering a stronger cognitive and physical connection to the task environment.

Collectively, these works highlight a consistent theme: SoSL, mostly driven by synchronized visual feedback and enhanced with haptic cues, is pivotal for effective and high-performance teleoperation. As recent research continues to demonstrate, the more immersive and responsive the feedback, the stronger the operator's SoSL becomes, ultimately enabling more intuitive and accurate teleoperation.

\subsection{Sense of Agency}
As the sense of agency (SoA) relies heavily on the trust of operator on that his/her actions will be mirrored by the surrogate, reinforcing this sense requires the design of adaptive and robust control algorithms capable of ensuring that the surrogate system accurately replicates human commands. From the perspective of control theory, accomplishment of such an objective relies on two criteria: \textit{stability} and \textit{transparency} \cite{hokayem2006bilateral} of the bilateral teleoperation system. The primary stability issue in bilateral teleoperation arises from time delays in the communication medium between the master and surrogate \cite{sirouspour2006model}. In addition, stability can be adversely affected by the system's interaction with both the human (active) and the environment (passive). Meanwhile, low transparency often stems from unknown uncertainties in the human/master and the surrogate/environment models. Though they may appear to be distinct objectives, stability and transparency are deeply interconnected. The central challenge in this field lies in achieving a high level of both, without compromising either \cite{feizi2022adaptive}. Over the past few decades, extensive research has been conducted to address these objectives \cite{ lawrence1993stability, hashtrudi2002transparency}. Therefore, enhancing the SoA fundamentally depends on achieving stable and transparent bilateral teleoperation control. This necessitates the development of robust and high-performance control strategies that can guarantee reliable and intuitive operation.

Many advanced control methods have been developed for electric master-surrogate systems with similar configurations \cite{chen2019rbf, guo2021adaptive, chen2019rbfnn, chen2019adaptive}, where joint-to-joint mapping is employed to close the control loop, limiting the applicability of methods to surrogates with dissimilar configurations. Dissimilar lightweight systems, still human-scale, with differing configurations and workspaces have also been investigated in the literature \cite{huang2024unified,slama2008experimental,li2025reinforcement,lu2024dynamic, ferraguti2020optimized, jorda2022local}. For example, in \cite{kastritsi2024passive}, a Force Dimension Sigma 7 haptic device is employed to teleoperate a 7-DoF KUKA robots by designing passivity-based admittance control scheme in the presence of time delay. The dissimilarity, furthermore, in the sense of cooperative teleoperation with single master and multi or bimanual surrogate systems has also been widely examined \cite{sirouspour2005modeling,sun2020single}. However, these approaches remain confined to systems within the human scale—where actuation dynamics, inertia, and control constraints are comparatively tractable. In contrast, the bilateral teleoperation of beyond-human-scale systems, such as HHMs, presents fundamentally different challenges. These include strong nonlinearities, significant time delays, and extreme motion/force scaling, which introduce severe sensorimotor disruption and control instability. Despite their growing industrial relevance, HHMs have been largely neglected in the teleoperation literature. Most prior work either assumes linear system behavior \cite{tafazoli2002impedance} or is limited to 1-DoF experimental setups \cite{banthia2017lyapunov, zarei2012lyapunov}, which are insufficient for the operational demands of real-world, high-DoF, nonlinear systems.

With increased attention to controlling hydraulically driven actuators using advanced control schemes and addressing accuracy and stability challenges, studies such as \cite{liang2024adaptive, xu2022extended, koivumaki2019energy, koivumaki2016stability} have paved the way for achieving stable and high-fidelity bilateral teleoperation of hydraulic manipulators. In \cite{lampinen2021force, lampinen2018full}, the bilateral teleoperation of 2-DoF HHMs with arbitrary motion-force scaling is examined, incorporating estimated human exogenous force and surrogate/environment interaction forces to ensure stability and transparency. To overcome the poor dynamics of HHMs, an integration of locally weighted intent prediction with a blended shared control method is proposed in \cite{luo2022human}. Furthermore, \cite{lampinen2018bilateral} investigates the bilateral teleoperation of 2-DoF HHMs in contact with both physical and virtual environments. The operational inefficiency of asymmetric master and hydraulically actuated surrogate is addressed in \cite{ding2022human} through a novel human-machine interface that enhances vision and incorporates auditory feedback. 

Consequently, most existing work on bilateral teleoperation focuses on symmetric or dissimilar human-scale systems with low DoF and electrically actuated robots. In contrast, teleoperation of large, beyond-human-scale systems—particularly HHMs—poses distinct challenges that remain largely unaddressed. These include nonlinear dynamics, extreme motion and force scaling, increased cognitive load, and sensorimotor mismatches. Ensuring transparency and stability under such conditions requires dedicated investigation for real-world deployment.

\subsection{Aims and Contributions}
\begin{figure*}
      \centering
      \subfloat[]{\includegraphics[width = 0.7\textwidth]{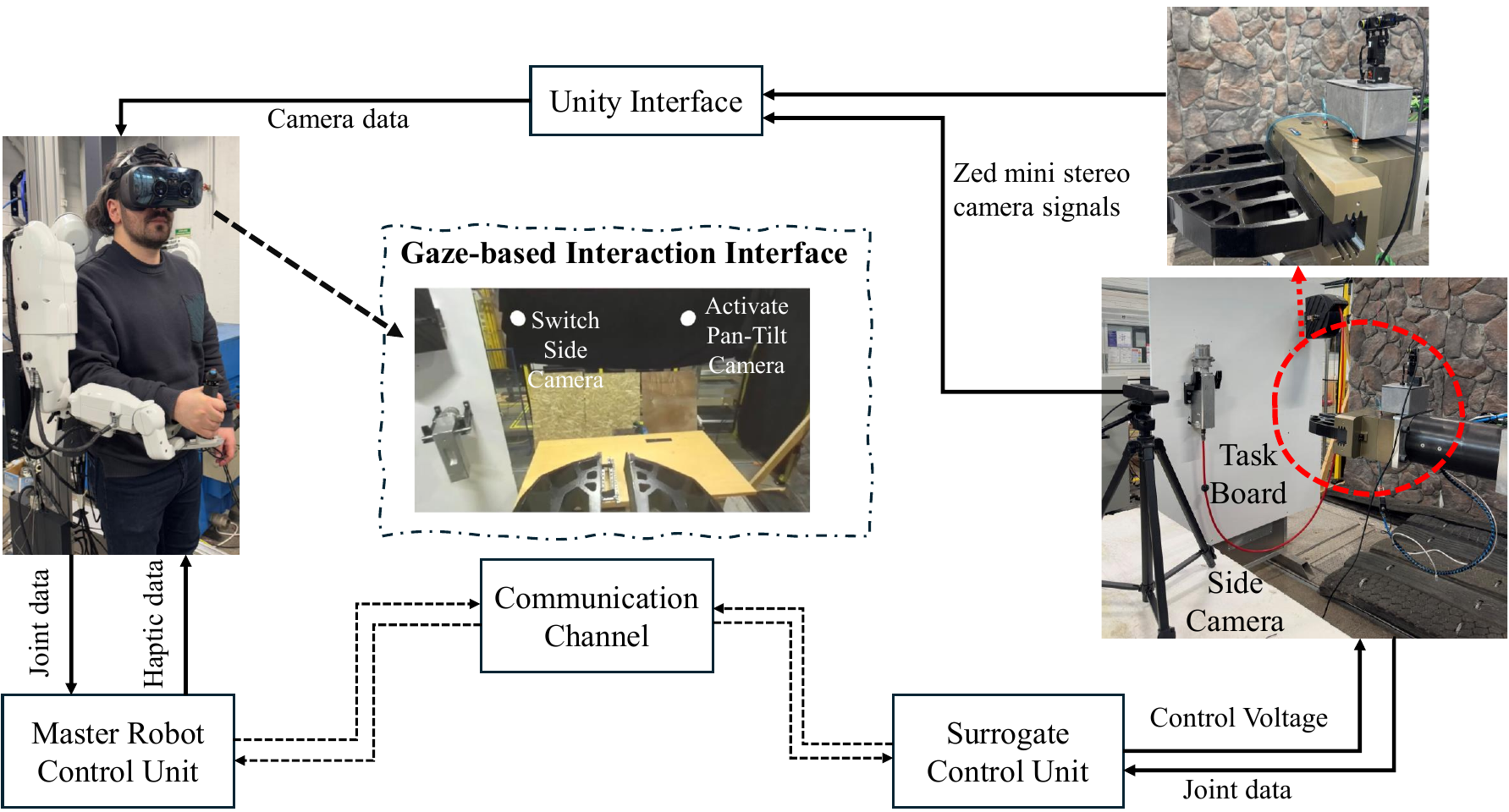}
      \centering
      \label{immersive_a}}
      \hfil
      \subfloat[]{\includegraphics[width = 0.3\textwidth]{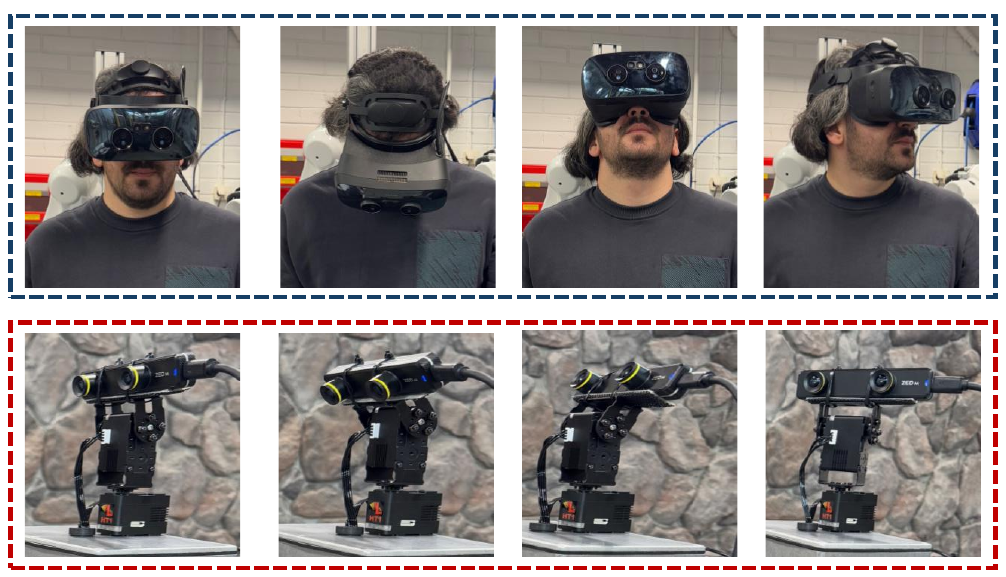}
      \centering
      \label{immersive_b}}
      \hfil
      \subfloat[]{\includegraphics[width = 0.4\textwidth]{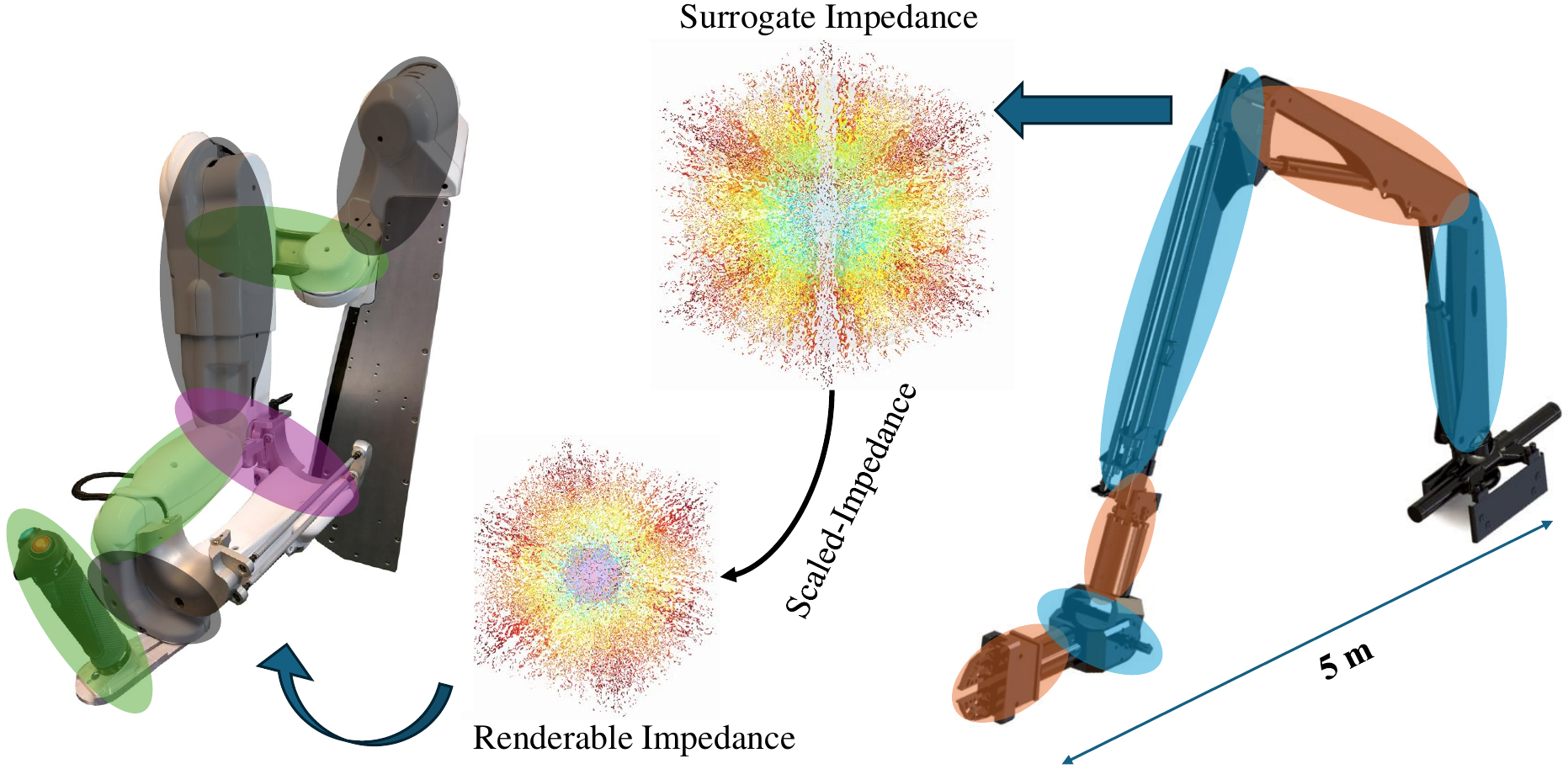}
      \centering
      \label{immersive_c}}
      \caption{Scheme of the VR-based immersive bilateral teleoperation system with egocentric head tracking system. (a) Overview of the general teleoperation framework, featuring control units, communication channels, and a gaze-based interaction interface. By utilizing the eye-tracking capability of the VR headset, the interface enables the operator to switch between modes based on gaze direction. The side camera further enhances situational awareness, especially in cases where the operator encounters difficulties during manipulation. (b) Pan-tilt mechanism of the remote camera, synchronized with the human operator’s egocentric head movements to maintain intuitive visual feedback. (c) Visualization of the rendered and distributed scaled impedance of the surrogate onto the human operator’s arm. The utilized haptic exoskeleton enables the operator to perceive both the dynamics of the surrogate and its interactions with the remote environment.}
      \label{immersive}
   \end{figure*}

Building on the previous discussion, this paper aims to significantly improve the accuracy and performance of bilateral teleoperation for large, heavy-duty robotic systems by enhancing immersion and reinforcing the operator’s SoE. As depicted in Fig. \ref{immersive_a}, we integrate a head-tracked pan-tilt camera system mounted on the surrogate, which synchronizes in real time with the operator’s natural egocentric head movements. This two-DoF active vision, as shown in Fig. \ref{immersive_b}, delivers immersive, first-person feedback that aligns with the operator’s viewpoint, reinforcing spatial presence and self-location \cite{wei2021multi}. Complementing the visual feedback, we employ ABLE, a 7-DoF haptic exoskeleton \cite{garrec2020design}, to deliver rich proprioceptive input (Fig. \ref{immersive_c}). This device distributes the scaled impedance of the surrogate along the operator’s arm and reflects its contact with remote environment, further strengthening the physical and cognitive link between operator and surrogate. In parallel, we design a robust, high-accuracy control framework capable of stabilizing the complex human-in-the-loop system under model uncertainties, time delays, and input nonlinearities. By maintaining stable and responsive behavior, this control architecture strengthens the operator’s SoA, fostering trust, precision, and control throughout the task. Therefore, the unique contributions of this work can be summarized as follows:

\begin{itemize}
    \item A VR-based visual feedback system with egocentric head-tracking capability is designed to deliver immersive, perspective-aligned visuals.The VR headset includes an eye-tracking feature that enables gaze-contingent resolution enhancement, increasing the level of immersion. To deepen the embodied experience, the haptic exoskeleton provides both contact force reflection and the distributed impedance of the surrogate. Together, these technologies reinforce the operator's SoE during task execution in a highly dissimilar teleoperation setup.
    
    \item A novel control strategy is developed to ensure the stability and transparency of the force-reflective bilateral teleoperation system. It accounts for the human-robot augmented dynamics, unknown model uncertainties, input nonlinearities, and arbitrary time delays. By embedding an interaction force estimation module for both the human/master and surrogate/environment sides, the proposed controller is well-suited for real-world deployment on heavy-duty manipulators, enabling intuitive, stable operation while enhancing the operator’s SoA.
    
    \item The proposed framework is deployed on a real-world heavy-duty teleoperation setup. Extensive experiments are conducted to evaluate the system’s robustness, transparency, and accuracy in both free-motion and contact-rich scenarios. Additionally, user studies assess the system’s usability and its effectiveness in enhancing SoE during task performance. 
\end{itemize}
Beyond improving task performance in challenging heavy-duty scenarios, the proposed system also serves as a rich platform for data collection and skill transfer. Its immersive, embodied setup is ideally suited for training learning-based models, including imitation learning, an area gaining traction particularly in the context of lightweight and humanoid systems \cite{cheng2024open, zhao2023learning}. This work thus lays a foundational step toward the full automation of beyond-human-scale robotic systems, combining human expertise with machine capability in a scalable, data-driven manner. 

The rest of the paper is organized as follows. Section \ref{Mathematics} explains the mathematical preliminaries. Section \ref{agency} elaborate on control design and stability analysis of master and surrogate systems. Section \ref{teleoperation} is dedicated to stability and transparency analysis of force-reflected bilateral teleoperation setup along with detail of VR-based visual feedback. The experimental results are provided in Section \ref{results}. Finally, Section \ref{conclusion} concludes this study.

\section{Mathematical Preliminaries}\label{Mathematics}
In this section, the mathematical foundation of the controller is explained along with the lemmas and assumptions employed in the rest of the paper.

\subsection{Virtual Decomposition Control}

\begin{figure*}[t]
\centering
\includegraphics[width=\textwidth]{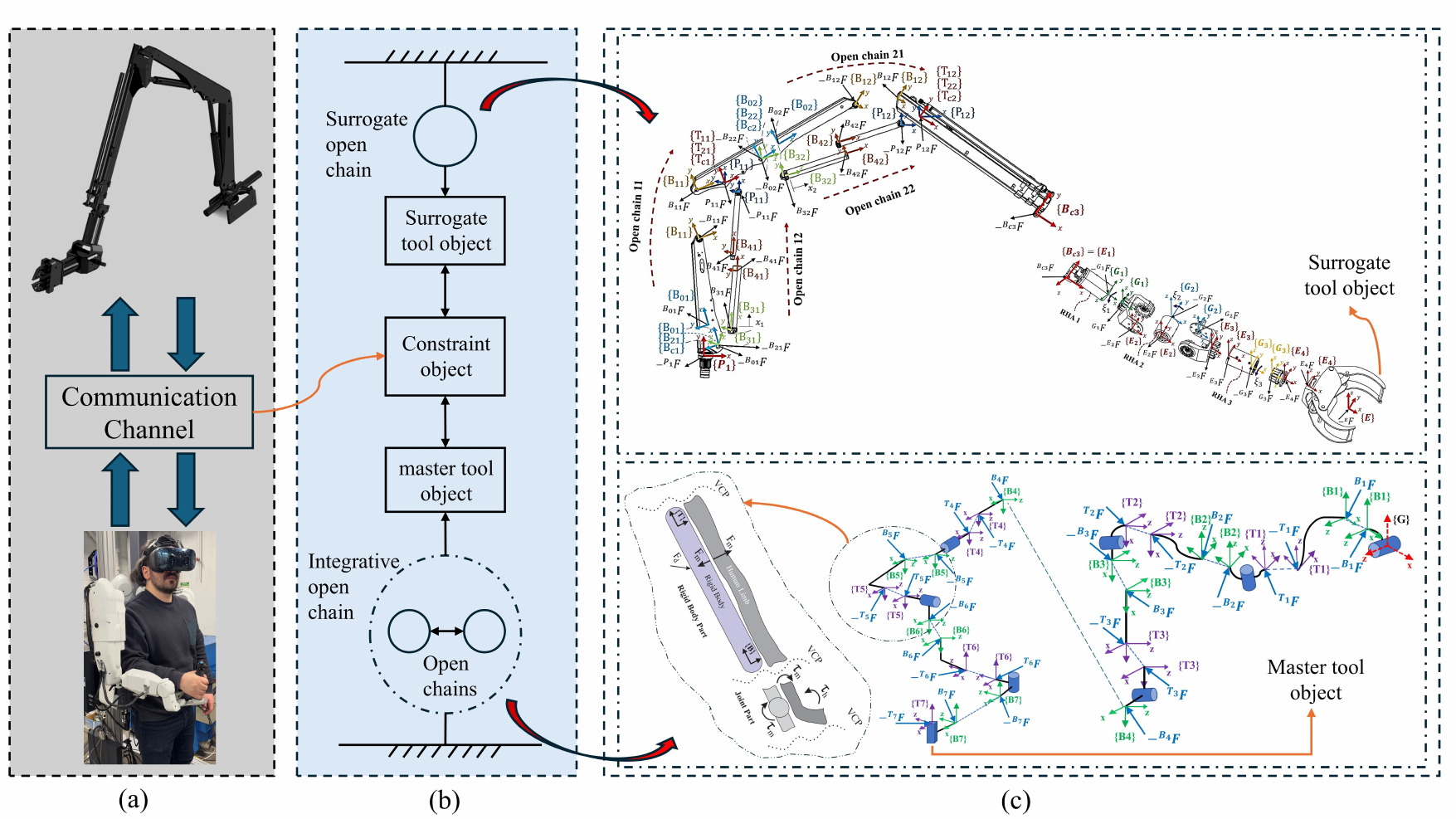}
\caption{ Proposed control method for robust and transparent force-reflected bilateral teleoperation. (a) Schematic of an immersive bilateral teleoperation system. (b) Graph-based representation of the proposed teleoperation control strategy. Open chains represent serial robotic manipulators; the "object" denotes the connective constraint linking two open chains; and the integrative open chain models the interconnected structure of the exoskeleton's serial links and the human arm. (c) Virtual decomposition of the master and surrogate systems. The integrative open chain highlights how the augmented human–robot model is seamlessly incorporated into the control framework. Details regarding the decomposition of the master robot and surrogate system can be found in \cite{hejrati2023physical} and \cite{hejrati2023orchestrated}, respectively.}
\label{VDC_scheme}
\end{figure*}
   
Virtual decomposition control (VDC) scheme is based on Plücker coordinate with compact formulation of system dynamics based on 6-D Plücker basis. In this coordinate system, \(\mathcal{M}^6\) denotes the space of spatial motions, while \(\mathcal{F}^6\) represents the space of spatial forces. Within these coordinates and spaces for a given frame \{A\} attached to a rigid body, the spatial velocity and force vectors can be defined as,

\begin{equation}
    ^AV = [\,^Av,\,^A\omega]^T \in \mathcal{M}^6, \quad \quad  ^AF = [\,^Af,\,^A\mathfrak{\tau}]^T \in \mathcal{F}^6,
\end{equation}
with \(^Av,\, ^A\omega ,\,^Af\), \(^A\mathfrak{\tau}\) being linear velocity, angular velocity, linear force, and moment, respectively, with respect to frame \{A\}. The spatial forces and velocity vectors can be transformed between frames as,

\begin{equation}
    ^BV =\, ^AU_B^T\,^AV, \quad \quad ^AF = \, ^AU_B\,^BF,
\end{equation}
with \(^AU_B\) being transformation matrix between two frames, defined as,
\begin{equation}
^{A}U_{B} = \begin{bmatrix}
^{A}R_{B} & \textbf{0}_{3\times3} \\
(^{A}r_{{A}{B}}\times)\, ^{A}R_{B} & ^{A}R_{B}
\end{bmatrix},
\label{eqn: A_U_B}
\end{equation}
where \(^{A}R_{B}\in \Re^{3\times3} \) is a rotation matrix between frame \{A\} and \{B\}, and (\(^{A}r_{{A}{B}}\times\)) is a skew-symmetric matrix operator defined as,
\begin{equation}\label{equ2}
^{A}r_{{A}{B}}\times = \begin{bmatrix}
0 & -r_z & r_y \\
r_z & 0 & -r_x\\
-r_y & r_x & 0
\end{bmatrix},
\end{equation}
where \(^{A}r_{{A}{B}} = [r_x,r_y,r_z]^T\) denotes a vector from the origin of frame \{A\} to the origin of frame \{B\}, expressed in \{A\}.

VDC scheme decomposes the system into rigid body and joint subsystems, shown in Fig. \ref{VDC_scheme}, for each of which the modeling, control, and stability analysis are performed separately at each local subsystem. Then the local stability is extended to the stability of entire system by means of virtual power flow (VPF) and virtual stability. Within this context, the dynamics of the rigid body is independent of the system and is expressed in 6-D vector as,
\begin{equation}
    M_A \dfrac{d}{dt}(^AV) + C_A(^A\omega)\,^AV + G_A(t) =\, ^AF^*,
    \label{RB dynamics}
\end{equation}
with \(M_A \in \Re^{6\times6}\), \(C_A \in \Re^{6\times6}\), and \(G_A\in \Re^{6}\) being spatial mass, centrifugal, gravitational matrices with \(^AF^* \in \Re^{6}\) being net spatial force applied to the rigid body \cite{hejrati2022decentralized}, all defined in Plücker coordinates.

\begin{prop}
    \cite{hejrati2022decentralized} The rigid body dynamics in (\ref{RB dynamics}) can be written in linear-in-parameter form as,
    \begin{equation}
        M_A \dfrac{d}{dt}(^AV) + C_A(^A\omega)\,^AV + G_A(t) = \Bar{Y}_A(^A\Dot{V}, \,^AV)\, \phi_A,
    \end{equation}
    with \(\Bar{Y}_A(^A\Dot{V}, \,^AV) \in \Re^{6\times10}\) being regression matrix and \(\phi_A(m_A,h_A, I_A) \in \Re^{10}\) being known inertial parameter vector. Additionally, \(m_A\), \(h_A\), and \(I_A\) are mass, first mass moment, and rotational inertia matrix, respectively.
\end{prop}

\begin{defka}
    \cite{zhu2010virtual} For the given frame \{A\}, the VPF is defined as the inner product of the spatial velocity vector error and the spatial force vector error,
    \begin{equation}\label{VPF}
    p_A = (^A\mathcal{V}_r-\,^A\mathcal{V})^T(^AF_r-\,^AF),
    \end{equation}
    where \(^A\mathcal{V}_r \in \mathcal{M}^6\) and \(^AF_r \in \mathcal{F}^6\) represent the required vectors of \(^A\mathcal{V} \in \mathcal{M}^6\) and \(^AF \in \mathcal{F}^6\), respectively.
    The inner product indicates the duality relationship between \(\mathcal{M}^6\) and \(\mathcal{F}^6\). 
    \label{Def pA}
\end{defka}

\begin{defka}
    A non-negative accompanying function \(\nu(t) \in \Re\) is a piecewise, differentiable function defined \(\forall \, t \in \Re^+\) with \(\nu(0) < \infty\) and \(\Dot{\nu}(t)\) exists almost everywhere.
    \label{nu def}
\end{defka}
\begin{lem}
    \cite{hejrati2023orchestrated} Consider a complex robot that is virtually decomposed into subsystems. Each subsystem is said to be virtually semi-globally uniformly ultimately bounded (SGUUB) with its non-negative accompanying function \(\nu(t)\) and its affiliated vector \(\Dot{\nu}(t)\), if and only if,
\begin{equation}\label{equ9}
{\Dot{{\nu}} \leq -\alpha_1 {\nu} + \alpha_{10}+p_A-p_C},
\end{equation}
with \(\alpha_1\) and \(\alpha_{10}\) being {positive constants} and \(p_A\) and \(p_C\) denoting the sum of VPFs in the sense of Definition \ref{Def pA}. This boundedness ensures robustness against unknown uncertainties, which is critical for real-world deployment.
\label{lemma SGUUB}
\end{lem}

Since the VDC is a velocity-based controller, the formulation of the required joint velocity plays a critical role. Its proper design enables the controller to handle a variety of tasks, such as free motion, in-contact interaction, and bilateral teleoperation. In this study, where the objective is to develop a controller for dissimilar bilateral teleoperation, the required end-effector velocities are first defined in~(\ref{Vmr}) and~(\ref{Vsr}). These velocities are then mapped to the corresponding joint space using the Jacobian matrices of the master and surrogate systems, resulting in the required joint velocities as shown in~(\ref{Dqr}) and~(\ref{Dtr}).

\subsection{Lemmas and Definitions}
\begin{lem}
    Thanks to their universal approximation capability, radial basis function neural networks (RBFNNs) can be employed to estimate any given continues function \(\mathbf{f}(t)\) represented in frame \{A\} as,
    \begin{equation}
        \hat{\mathbf{f}}_A(t) =\, {^{ {A}}}\hat{W}^T\, \psi(\chi_A) + {^{ {A}}}\hat{\epsilon},
    \end{equation}
    with \(\chi_A\) being the input vector, \({^{ {A}}}\hat{W}\) being the estimated weights, \(\psi(.)\) being Gaussian basis function, and \({^{ {A}}}\hat{\epsilon}\) being estimation error of the RBFNNs. The adaptation laws for estimation of weights and bias of the network can be considered as follows \cite{hejrati2023orchestrated},
    \begin{equation}
        {^{ {A}}\Dot{\hat{W}} = {\Pi}\,\left(\psi(\chi_{ {A}})\,(^{ {A}}{ V}_r-\,^{ {A}}{ V})^T-\, \tau_0\,{^{ {A}}\hat{W}}\right)},
    \end{equation}
    \begin{equation}
        {{^{ {A}}}\Dot{{\epsilon}} = {\pi} \left (({^{ {A}}{ V}_r} - {^{ {A}}{ V}})- \pi_0 {^{ {A}}}{\epsilon}\right),}
        \label{eps adapt}
    \end{equation}
    where \(\Pi, \tau_0, \pi\), \(\pi_0\) are positive constants. Under the optimal weights of RBFNNs as,
    \begin{equation}
        W^* = \arg \min_{\hat{W} \in\, \Xi_N}  \lbrace \sup_{\chi \in\, \Xi_T}|\hat{\mathbf{f}}(\chi|\hat{W})-{\mathbf{f}}(\chi)| \rbrace,
    \end{equation}
    where \(\Xi_N = \lbrace \hat{W}|\lVert \hat{W} \rVert \leq \kappa \rbrace\) is a valid set of vectors with \(\kappa\) being a design value, \(\Xi_T\) is an allowable set of the state vectors, there is,
    \begin{equation}
        |\hat{\mathbf{f}}(\chi)-\,^A{W^*}^T\, \psi(\chi_A)|=\,|\epsilon^*(\chi_A)|\leq\Bar{\epsilon},
    \end{equation}
    where \(\Bar{\epsilon}\) is an unknown bound parameters.
    \label{RBFNNs}
\end{lem}

\begin{lem}  
\cite{tee2009barrier}  
Let \(k_b > 0\) be a positive constant, and define the sets \(\mathfrak{Z} := \lbrace \mathfrak{z} \in \Re : -k_b < \mathfrak{z} < k_b \rbrace \subset \Re\) and \(\mathcal{N} := \Re^\mathfrak{l} \times \mathfrak{Z} \subset \Re^{\mathfrak{l}+1}\), which are open. Consider the dynamical system:  
\begin{equation}
    \Dot{\eta} = \hbar(t, \eta),
\end{equation}
where \(\eta := [\omega, \mathfrak{z}]^\top \in \mathcal{N}\), and \(\hbar : \Re_+ \times \mathcal{N} \to \Re^{\mathfrak{l}+1}\) is piecewise continuous in \(t\) and locally Lipschitz in \(\eta\), uniformly in \(t\), on \(\Re_+ \times \mathcal{N}\). Suppose there exist functions \(\mathfrak{U} : \Re^\mathfrak{l} \to \Re_+\) and \(\mathfrak{V} : \mathfrak{Z} \to \Re_+\), both continuously differentiable and positive definite in their respective domains, such that:  
\begin{equation}
\mathfrak{V}(\mathfrak{z}) \to \infty \quad \text{as} \quad \mathfrak{z} \to -k_b \quad \text{or} \quad \mathfrak{z} \to k_b,
\end{equation}
\begin{equation}
\delta_7(\lVert\omega\rVert) \leq \mathfrak{U}(\omega) \leq \delta_8(\lVert\omega\rVert),
\end{equation} 
where \(\delta_7\) and \(\delta_8\) are class \(\mathcal{K}_\infty\) functions. Let \(\mathrm{V}(\eta) := \mathfrak{V}(\mathfrak{z}) + \mathfrak{U}(\omega)\), and assume \(\mathfrak{z}(0) \in (-k_b, k_b)\). If the following inequality holds:  
\begin{equation}
\Dot{\mathrm{V}} = \frac{\partial \mathrm{V}}{\partial \eta} \hbar \leq 0,
\end{equation} 
then \(\mathfrak{z}(t)\) remains in the open set \((-k_b, k_b)\) for all \(t \in [0, \infty)\).  
\label{BLF}
\end{lem}  

\begin{defka} \label{121 map}
    For any unique representation of inertial parameter vector of rigid body \(\phi_A\), there is one-to-one map as,
    \begin{equation}
    \mathcal{N}(\phi_{A})= \mathcal{L}_{A} = \begin{bmatrix}
        0.5tr(\Bar{I}).\textbf{1}-I & h \\
        h^T & m
        \end{bmatrix},
    \end{equation}
    \begin{equation}
    \mathcal{N}^{-1}(\mathcal{L}_{A}) = \phi_{A}(m,h,tr(\Sigma).\textbf{1}-\Sigma),
    \end{equation}
    where \(\Sigma = 0.5tr(I)-I\) and \(tr(.)\) is the Trace operator of a matrix. 
\end{defka}
\begin{defka}
    {For a given \mbox{\(\mathcal{L}_{A}\)} with its estimation \mbox{\(\mathcal{\hat{L}}_{A}\)}, the natural adaptation law (NAL) can be derived as,}
\begin{equation}
        {\Dot{\hat{\mathcal{L}}}_A = \frac{1}{\gamma}\, \hat{\mathcal{L}}_A\,\left(\mathcal{S}_A-\gamma_0\,\hat{\mathcal{L}}_A\right)\, \hat{\mathcal{L}}_A},
        \label{L adapt}
\end{equation}
{with \mbox{\(\gamma >0\)} being the adaptation gain for all the rigid bodies of the system, and \mbox{\(\gamma_0>0\)} being a small positive gain. Additionally, \mbox{\(\mathcal{S}_A\)} is a unique symmetric matrix defined in \mbox{\cite{hejrati2023physical}}}.

\subsection{{Notation}}
{The following notation conventions are employed throughout the paper. For a general signal \mbox{\(\mathfrak{n}\)}:}
{\mbox{\(\boldsymbol{\mathfrak{n}}\)} denotes the filtered form of \mbox{\(\mathfrak{n}\)}, computed via,}
    \begin{equation}
            \Dot{\boldsymbol{\mathfrak{n}}} + \mathcal{C} \boldsymbol{\mathfrak{n}} = \mathcal{C} \mathfrak{n},
    \end{equation}
    {where \mbox{\(\mathcal{C} > 0\)} is a scalar for scalar signals or a positive-definite matrix for vector signals.}
{\mbox{\(\hat{\mathfrak{n}}\)} represents the estimated value of \mbox{\(\mathfrak{n}\)}.}
{\mbox{\(\Tilde{\mathfrak{n}} = \hat{\mathfrak{n}} - \mathfrak{n}\)} denotes the estimation error.}
{\mbox{\({}^A\mathfrak{n}\)} refers to \mbox{\(\mathfrak{n}\)} expressed in coordinate frame \mbox{\(\{A\}\)}.}
{These forms are used consistently to represent signals in filtering, estimation, and frame-dependent computations across the master and surrogate systems.}

\section{{Sense of Agency through Control}}
\label{agency}
{This section presents the control design for both the master and surrogate robots, which interact with the human operator and the environment, respectively.}

\subsection{Augmented Human/Master Dynamics and Control} \label{Master side}
The master robot utilized in this study is a 7-DoF haptic exoskeleton, shown in Fig. \ref{Exo}, which serves as both an exoskeleton and a haptic display. The objective of this section is to design a control action that not only renders the scaled impedance of the surrogate and its interaction with the environment, but also guarantees the safety of the human in the loop. The details of the controller design are presented in the following.

The decentralized human-robot augmented model, shown in Fig. \ref{VDC_scheme}c, proposed by \cite{hejrati2023physical} can be written as follows,
\begin{equation}
    \mathcal{M}_B\frac{d}{dt}(^{B_i}\mathcal{V})+\,\mathcal{C}_{B_i}\,^{B_i}\mathcal{V}+\,\mathcal{G}_{B_i} =\, ^{B_i}\mathcal{F}^*-\Delta_{D_i},
    \label{HR body}
\end{equation}
\begin{equation}
    \mathcal{I}_i\Ddot{q}_i = \tau^*_i-\tau_{h_i}-\Delta_J,
    \label{HR Joint}
\end{equation}
\begin{equation}
    \tau_i = \tau^*_i + \tau_{ai},
    \label{tau}
\end{equation}
{where \mbox{\(\mathcal{M}_B = M_R + M_h \in \Re^{6\times6}\)}, \mbox{\(\mathcal{C}_B = C_R + C_h \in \Re^{6\times6}\)}, and \mbox{\(\mathcal{G}_B = G_R + G_h \in \Re^{6\times6}\)}} are augmented matrices in the form of (\ref{RB dynamics}), {\mbox{\(\mathcal{I} = I_R +I_h \in \Re^6\)}} being augmented rotational inertia of the joint, \(\tau_h\) being exogenous human torque (EHT), \(\Delta_D \in \Re^{6}\) and \(\Delta_J \Re^{6}\) encompassing unmodeled dynamics and compound input nonlinearity for rigid body and joint subsystems, respectively, with {\mbox{\((.)_R\)}} standing for robot and \((.)_h\) representing human-related terms.

\begin{rmk}
    Equation (\ref{tau}) presents the recomposed dynamics of the decomposed system. The first term on the right-hand side of (\ref{tau}) represents the torque responsible for generating joint-level actions, while the second term, \(\tau_{a_i} = \sigma^T\, ^{B_i}\mathcal{F}\) where \(^{B_i}\mathcal{F}\) defined in (\ref{equ22}), corresponds to the forces arising from the motion of the rigid body. It worth to note that the screw vector \(\sigma\) maps the spatial force \(^{B_i}\mathcal{F}\) from spatial forces space to the joint space.
\end{rmk}

Being grounded in the Newton-Euler recursive method, VDC necessitates the computation of forward spatial velocity and acceleration, as well as backward spatial force propagation. These steps are fundamental for designing the control law in Plücker coordinates. In the following, the computation process is detailed for the 7-DoF haptic exoskeleton.

\subsubsection{Kinematics and Dynamics}
Considering the framing convention depicted in Fig. \ref{Exo_a}, the spatial velocity vectors for the \(i\)-th link of the decentralized system can be derived starting from the base frame as follows:  

\begin{equation}\label{equ17}
    ^{B_i}\mathcal{V} =\, ^{T_{i-1}}U_{B_i}^T\, ^{T_{i-1}}\mathcal{V} + \sigma_i \Dot{q}_i, 
\end{equation}  
\begin{equation}\label{equ18}
    ^{T_i}\mathcal{V} =\, ^{B_i}U_{T_i}^T\, ^{B_i}\mathcal{V},
\end{equation}  
for \(i = 1, \ldots, 7\), with \(\Dot{q}_i\) being joint angular velocity. Once the spatial velocity and, consequently, the spatial acceleration are computed, the spatial forces can be determined as follows:  
\begin{equation}\label{equ22}
    ^{B_j}\mathcal{F} =\, ^{B_j}U_{T_j}\, ^{T_j}\mathcal{F} +\, ^{B_j}\mathcal{F}^*, 
\end{equation}  
\begin{equation}\label{equ23}
    ^{T_{j-1}}\mathcal{F} =\, ^{T_{j-1}}U_{B_j}\, ^{B_j}\mathcal{F}, 
\end{equation}  
for \(j = 7, \ldots, 1\), where \(^{B_j}\mathcal{F}^*\) represents the net spatial force applied to frame \(\{B\}\) as defined in (\ref{HR body}). Fig. \ref{Exo_b} illustrates the frames and spatial forces applied to the \(i^{th}\) decomposed subsystem. With the dynamics defined in (\ref{HR body})–(\ref{equ23}), the corresponding control action can now be designed, as explained in the following sections.

\begin{figure}[pos=t]
      \centering
      \subfloat[]{\includegraphics[width = 0.2\textwidth]{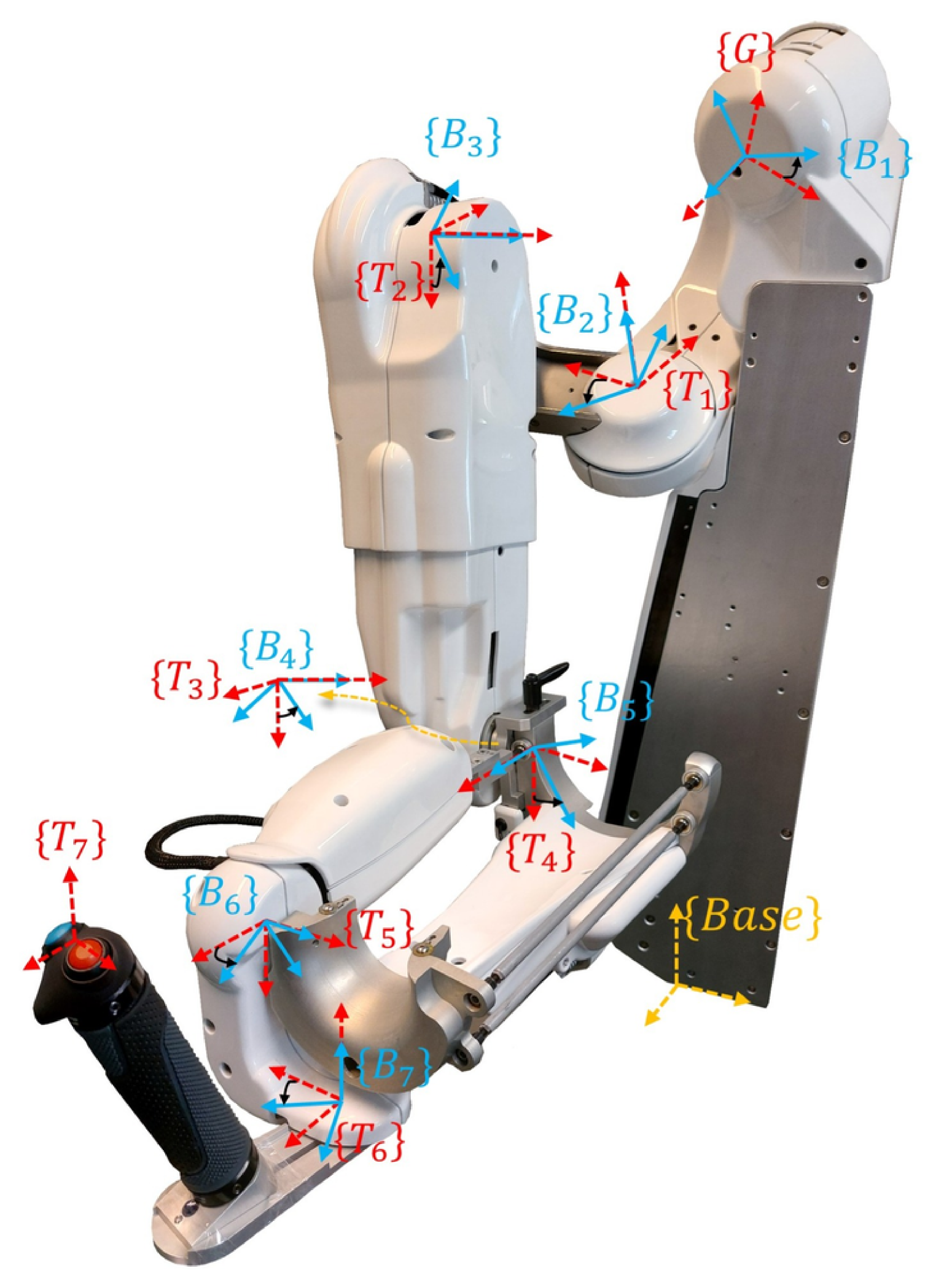}
      \centering
      \label{Exo_a}}
      \hfil
      \subfloat[]{\includegraphics[width = 0.3\textwidth]{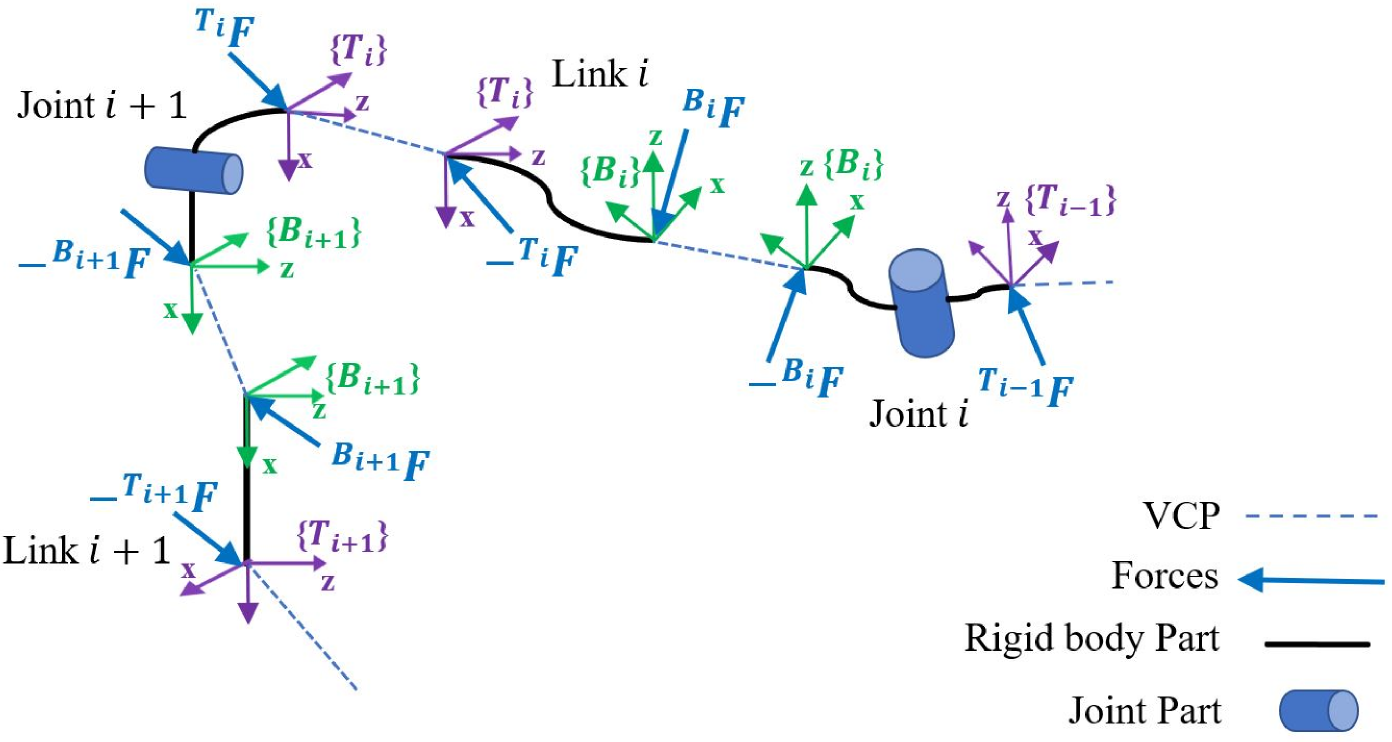}
      \centering
      \label{Exo_b}}
      \caption{Decomposition of haptic exoskeleton using the VCP concept. a) Framing convention in VDC context, b) \(i^{th}\) subsystem of decomposed model}
      \label{Exo}
   \end{figure}

\subsubsection{Control design}
In order to establish the force-reflected bilateral teleoperation, the required end-effector velocity of master robot can be defined as,
\begin{equation}
    V_{mr} = V_{md} - \mathcal{A}\boldsymbol{F}_m,
    \label{Vmr}
\end{equation}
where \(V_{md} \in \Re^{6}\) is defined in~(\ref{Vmd}), and \(\boldsymbol{F}_m \in \Re^{6}\) denotes the estimated interaction force inserted from exoskeleton to the human arm, computed using the method proposed in~\cite{hejrati2024impact}.
Then, using the pseudo-inverse of Jacobian matrix \(\mathcal{J}_m \in \mathcal{R}^{6\times7}\), the joint level required velocity can be derived as,
\begin{equation}
    \Dot{q}_r = \mathcal{J}_m^\dagger\, V_{mr}.
    \label{Dqr}
\end{equation}
The velocity term in (\ref{Dqr}) represents the motion that must be generated by the actuators of the master robot to fulfill the objectives of teleoperation. With this notion, the required spatial velocity of the rigid bodies can be computed by substituting (\ref{Dqr}) into (\ref{equ17})–(\ref{equ18}) as follows:
\begin{equation}\label{equ19}
     ^{B_i}\mathcal{V}_r =\, ^{T_{i-1}}U_{B_i}^T\, ^{T_{i-1}}\mathcal{V}_r + \sigma_i \Dot{q}_{ri},
\end{equation}
\begin{equation}\label{equ20}
     ^{T_i}\mathcal{V}_r =\, ^{B_i}U_{T_i}^T\, ^{B_i}\mathcal{V}_r.
\end{equation}
In the context of VDC, the forces that generate the required spatial velocities are referred to as required spatial forces. These forces correspond to those that would produce the required velocity described in (\ref{equ19}) if applied to the rigid body. Consequently, the required spatial forces can be expressed as follows:
\begin{equation}\label{equ24}
    ^{B_j}\mathcal{F}_r =\, ^{B_j}U_{T_j}\, ^{T_j}\mathcal{F}_r +\, ^{B_j}\mathcal{F}_r^*,
\end{equation}
\begin{equation}\label{equ25}
    ^{T_{j-1}}\mathcal{F}_r =\, ^{T_{j-1}}U_{B_j}\, ^{B_j}\mathcal{F}_r,
\end{equation}
with \(^{B_j}\mathcal{F}_r^*\) being the required net spatial force of the rigid body defined in (\ref{F*r}). Finally, the unified required joint torque can be computed in the sense of (\ref{tau}) as,
\begin{equation}\label{equ26}
    \tau_{ri} = \tau^*_{ri} + \tau_{ari},
\end{equation}
with \(\tau_{ari} = \sigma_i^T\, ^{B_i}\mathcal{F}_r\), and \(\tau^*_{ri}\) defined in (\ref{tau*}). Addressing the unknown uncertainties and dynamics of the rigid bodies and joints, the control signal in (\ref{equ26}) achieves the objectives of the teleoperation system by generating the required joint velocity described in (\ref{Dqr}).

\begin{thm}
    For the decentralized and augmented model of pHRI described in (\ref{HR body})–(\ref{tau}), along with the spatial velocity and force computations in (\ref{equ17})–(\ref{equ23}) and (\ref{equ19})–(\ref{equ25}), the decentralized control actions can be designed as follows:
\begin{equation}
\begin{split}
    ^{B_i}\mathcal{F}^*_r &=\,\mathcal{K}_{Di}(\,^{B_i}\mathcal{V}_r-\,^{B_i}\mathcal{V}) + \hat{W}_{Di}^T\Psi(\chi_{Di})+\hat{\varepsilon}_{Di} \\
    & +Y_{B_i}\hat{\phi}_{B_i},
\end{split}
\label{F*r}
\end{equation}
\begin{equation}
\begin{split}
    \tau^*_{ri} &= k_{di}(\Dot{q}_{ri}-\,\Dot{q}_i) +Y_{ai}\hat{\phi}_{ai} + \hat{W}_{Ji}^T\Psi(\chi_{Ji})\\
    &+\hat{\varepsilon}_{Ji}+\frac{e_{ai}+c_1\Dot{e}^\dagger e^2_{ai}}{k^2_{bi}-e^2_{ai}},
\end{split}
\label{tau*}
\end{equation}
which ensures the SGUUB of all signals in the closed loop system. Additionally, the followings hold,
\begin{equation}
    e_{ai} \in \Omega_e,
\end{equation}
\begin{equation}
    \rho_m \defequal V_{mr}-V_m =  V_{md}-V_m-\mathcal{A}\boldsymbol{F}_m \in \Omega_m,
    \label{rho_m}
\end{equation}
\begin{equation}
    \Omega_m \defequal \lbrace V_{mr}-V_m \in \Re^6|\, ||V_{mr}-V_m||\leq \mathcal{D}_m \rbrace,
\end{equation}
\begin{equation}
    \Omega_e \defequal \lbrace e_{ai} \in \Re|\, |q_{ri}-q_i|\leq \mathcal{D}_1, \|\Dot{q}_{ri}-\Dot{q}_i\| \leq \mathcal{D}_2 \rbrace.
\end{equation}
\label{Thm Exo}
\end{thm}
\begin{pf}
Defining non-negative accompanying function as,
    \begin{equation}
        \begin{split}
            \nu_m(t) &= \sum_{i=1}^{7} (\nu_i(t)+\nu_{ai}(t)),\\
        \end{split}
        \label{nu}
    \end{equation}
    \begin{equation}\label{equ35}
    \begin{split}
    \nu_i(t) &=\, \frac{1}{2} (^{B_i}\mathcal{V}_r-\,^{B_i}\mathcal{V})^T\, \mathcal{M}_{B_i}\,(^{B_i}\mathcal{V}_r-\,^{B_i}\mathcal{V})\\ 
    &+\gamma_1\mathcal{D}_F(\mathcal{L}_{B_i}\rVert \hat{\mathcal{L}}_{B_i}) + \frac{1}{2}tr(\Tilde{W}_{Di}^T\,\Gamma_{B_i}^{-1}\Tilde{W}_{Di}) 
    \\
    &+  \frac{1}{2\gamma_{2i}} \Tilde{\varepsilon}_{Di}^T\,\Tilde{\varepsilon}_{Di}.
    \end{split}
    \end{equation}
    \begin{equation}\label{equ46}
    \begin{split}
        \nu_{ai}(t) &=\, \frac{1}{2}\mathcal{I}_i (\Dot{q}_{ri}-\,\Dot{q}_i)^2+\frac{1}{2}\log{\frac{k^2_{bi}}{k^2_{bi}-e^2_{ai}}}\\
        &+\zeta\mathcal{D}_F(\mathcal{L}_{ai}\rVert \hat{\mathcal{L}}_{ai}) + \frac{1}{2\beta_{2i}} \Tilde{\varepsilon}_{Ji}^2 + \frac{1}{2\beta_{1i}}\Tilde{W}_{Ji}^T\,\Tilde{W}_{ji},
    \end{split}
    \end{equation}
    and taking its time derivative, replacing adaptation functions from Lemma \ref{RBFNNs} and (\ref{L adapt}), the local controllers in (\ref{F*r}) and (\ref{tau*}) along with employing Young's inequality as \(-tr(\hat{\mathcal{L}}_{B_i}\Tilde{\mathcal{L}}_{B_i}) \leq -0.5tr(\Tilde{\mathcal{L}}_{B_i}\Tilde{\mathcal{L}}_{B_i})+ 0.5tr({\mathcal{L}}_{B_i}{\mathcal{L}}_{B_i})\), \(-tr(\hat{W}^T_{D_i}\Tilde{W}_{D_i}) \leq -0.5tr(\Tilde{W}^T_{D_i}\Tilde{W}_{D_i})+ 0.5tr({W}^T_{D_i}{W}_{D_i})\), \(-\hat{\varepsilon}^T_{Di}\Tilde{\varepsilon}_{Di} \leq - 0.5\Tilde{\varepsilon}^T_{Di}\Tilde{\varepsilon}_{Di}+0.5{\varepsilon}^T_{Di}{\varepsilon}_{Di}\), \(-\hat{\varepsilon}^T_{Ji}\Tilde{\varepsilon}_{Ji} \leq - 0.5\Tilde{\varepsilon}^T_{Ji}\Tilde{\varepsilon}_{Ji}+0.5{\varepsilon}^T_{Ji}{\varepsilon}_{Ji}\), we have,
\begin{equation}
\begin{split}
        \Dot{\nu}_m(t) &\leq -\sum_{i = 1}^{n}[(\,^{B_i}\mathcal{V}_r-\,^{B_i}\mathcal{V})^T\,\mathcal{K}_{Di}\,(\,^{B_i}\mathcal{V}_r-\,^{B_i}\mathcal{V})\\
        & + k_{di}(\Dot{q}_{ri}-\,\Dot{q}_i)^2 +0.5\gamma_{0m}tr(\Tilde{\mathcal{L}}_{B_i}\Tilde{\mathcal{L}}_{B_i})\\
        &+0.5\tau_{0m}tr(\Tilde{W}^T_{Di}\Tilde{W}_{Di})+\pi_{0m}\Tilde{\varepsilon}^T_{Di}\Tilde{\varepsilon}_{Di}+\pi_{0m}\Tilde{\varepsilon}^T_{Ji}\Tilde{\varepsilon}_{Ji}\\
        & +0.5\gamma_{0m}tr(\Tilde{\mathcal{L}}_{ai}\Tilde{\mathcal{L}}_{ai})+0.5\tau_{0m}tr(\Tilde{W}^T_{Ji}\Tilde{W}_{Ji})\\
        & +c_1 \dfrac{e^2_{ai}}{k^2_{bi}-e^2_{ai}}] + \sum_{i = 1}^{n}[0.5\gamma_{0m}tr({\mathcal{L}}_{B_i}{\mathcal{L}}_{B_i}) \\
        &+ 0.5\tau_{0m}tr({W}^T_{Di}{W}_{Di})+\pi_{0m}{\varepsilon}^T_{Di}{\varepsilon}_{Di}+\pi_{0m}{\varepsilon}^T_{Ji}{\varepsilon}_{Ji}\\
        & +0.5\gamma_{0m}tr({\mathcal{L}}_{ai}{\mathcal{L}}_{ai})+0.5\tau_{0m}tr({W}^T_{Ji}{W}_{Ji})]\\
        & \leq - \vartheta \nu_m(t)+\vartheta_0 - p_{T_7},
\end{split}
\label{Dvm}
\end{equation}
where,
\begin{equation}
    \begin{split}
        \vartheta_0 &=  0.5\gamma_{0m}tr({\mathcal{L}}_{B_i}{\mathcal{L}}_{B_i}) + 0.5\tau_{0m}tr({W}_{Di}{W}_{Di})+\pi_{0m}{\varepsilon}_{Di}{\varepsilon}_{Di}\\
        &+\pi_{0m}{\varepsilon}_{Ji}{\varepsilon}_{Ji} +0.5\gamma_{0m}tr({\mathcal{L}}_{ai}{\mathcal{L}}_{ai})+0.5\tau_{0m}tr({W}_{Ji}{W}_{Ji}), 
    \end{split}
\end{equation}
\begin{equation}
    \vartheta = \vartheta_{min}/\vartheta_{max},
\end{equation}
\begin{equation}
    \begin{split}
        \vartheta_{min} = \min(\min(k_{di}), \min(\mathcal{K}_{Di}),\gamma_{0m},\tau_{0m},c_1,\pi_{0m}),
    \end{split}
\end{equation}
and,
\begin{equation}
\begin{split}
        \vartheta_{max} &= \max(\max(M_{B_i}),\max(\mathcal{K}_{D_i}),\gamma_{1m},\max(\Gamma_{B_i}),\\
        &\max(\gamma^{-1}_{2i}),\max(\mathcal{I}_i),\max(\beta^{-1}_{2i}),\max(\beta^{-1}_{1i}),\zeta,1),
\end{split}
\end{equation}
where the following inequality is used,
\begin{equation}
    \log{\frac{k^2_{bi}}{k^2_{bi}-e^2_{ai}}} \leq \frac{e^2_{ai}}{k^2_{bi}-e^2_{ai}},
\end{equation}
According to (\ref{Dvm}) all the VPFs cancel out each other as shown in Appendix B of \cite{hejrati2023physical} except \(p_{T_7}\), which indicates the virtual power flow between human hand and the robot. Then, by integrating both side of (\ref{Dvm}), one can obtain,
\begin{equation}
    \begin{split}
        \nu_m(t)\leq \nu(0)e^{-\vartheta\,t} + \dfrac{\vartheta_0}{\vartheta} (1-e^{-\vartheta\,t}) + \varrho_m,
    \end{split}
    \label{vm}
\end{equation}
with \( \varrho_m\) defined in Appendix \ref{p_T_7}. According to (\ref{vm}) and Appendix \ref{p_T_7}, the maximum and steady-state value of \(\nu_m\) can be derived as, \(\nu_m(t) \leq \nu_{max} = \nu(0) + \vartheta_0/{\vartheta}+\varrho_m\) and \(\lim_{t \to \infty} \nu_m (t) \leq \vartheta_0/{\vartheta}+\varrho_{0m}\), with \(\varrho_{0m}\) defined in Appendix \ref{p_T_7}. Then, the following holds from (\ref{nu}),
\begin{equation}
    \frac{1}{2}\log{\frac{k^2_{bi}}{k^2_{bi}-e^2_{ai}}} \leq \nu_m (t) \leq \nu_{max}.
    \label{log}
\end{equation}
Taking the exponential on both sides of (\ref{log}) and rearranging the terms, we can show that,
\begin{equation}
    |e_{ai}| \leq k_{bi} \sqrt{1-e^{-2\nu_{max}}} = \mathcal{D}_1.
\end{equation}
In the same way, it can be shown that,
\begin{equation}
    \|\Dot{q}_{ri}-\Dot{q}_i\| \leq \sqrt{\dfrac{2\nu_{max}}{\lambda_{min}(\mathcal{I}_i)}} = \mathcal{D}_2,
\end{equation}
\begin{equation}
    ||V_{mr}-V_m|| \leq \alpha_m \sqrt{\dfrac{2\nu_{max}}{\lambda_{min}(M_{B_7})}} = \mathcal{D}_m,
\end{equation}
with \(||^{B_7}U_{T_7}|| \leq \alpha_m\). Additionally, considering Lemma \ref{BLF} and (\ref{vm}), we have \(|e_{ai}| \leq k_{bi}\). Then from \(q_{i} = q_{ri} - e_{ai} \) and \(|q_{ri}| \leq k_{ri}\), we can conclude that \(|q_i| \leq k_{ci} = k_{ri} + k_{bi}\) if \(e_{ai}(0) \leq k_{bi}\), ensuring that state-constraints will not be violated.

\end{pf}

\subsection{Surrogate/Environment Dynamics and Control}\label{slave side}
The surrogate in this study is a 6-DoF HHM with a 3-DoF anthropomorphic arm and a spherical wrist, shown in Fig. \ref{Hiab}, encompassing the complexities arising from fluid dynamics and the nonlinearities of a high-DoF system. The primary objective of the surrogate is to accurately follow the motion and force commands from the master robot while ensuring stable contact with an unknown environment. Furthermore, since the use of force sensors in HHMs is not always a feasible solution, a force-sensorless control law is designed to address this issue effectively.

For doing so, the required end-effector velocity of the surrogate can be defined as,
\begin{equation}
    V_{sr} = V_{sd} - \mathcal{A}{\boldsymbol{F}}_s,
    \label{Vsr}
\end{equation}
with \(V_{sd} \in \Re^6\) defined in (\ref{Vsd}) and \({\boldsymbol{F}}_s \in \Re^6\) denotes the estimated interaction force between surrogate end-effector and remote environment, computed using the method proposed in~\cite{hejrati2024impact}. Then, using the damped least square (DLS) inverse of Jacobian matrix \(\mathcal{J}_s \in \mathcal{R}^{6\times6}\), one can obtain the required joint velocity as follows,
\begin{equation}
    \Dot{\theta}_r = \lbrace(\mathcal{J}_s^T\,\mathcal{J}_s+\lambda I)^{-1}\,\mathcal{J}_s^T\rbrace\, V_{sr},
    \label{Dtr}
\end{equation}
with \(\lambda \in \mathcal{R}\) being damping factor. The use of DLS method tackles the gimbal lock effect, which is a common phenomenon in spherical wrists. To enhance readability and avoid redundancy, only the tool of the surrogate is considered to demonstrate the control design in this study. The comprehensive analysis of the entire system, for both free-motion and in-contact tasks, is elaborated in \cite{hejrati2023orchestrated} and \cite{hejrati2024impact}. Therefore, as shown in Fig. \ref{Hiab}, it is assumed that the velocity computations and force backpropagations are performed up to frame \(\{G_3\}\) for both \(\Dot{\theta}\) and \(\Dot{\theta}_r\) in (\ref{Dtr}), as described in \cite{hejrati2023orchestrated}.
 
At the time of contact, the following hold
\begin{equation}\label{TF}
    {^T}F = N_c\ f_{e},
\end{equation}
where,
\begin{equation}
    f_{e} = M_e \Dot{V}_{s} + D_e V_s + K_e \mathbf{x}_e,
    \label{f_e}
\end{equation}
with \(\mathbf{x}_e\) being the deformation of the environment, \(M_e\), \(D_e\), and \(K_e\) being the environmental mass, damping, and stiffness matrices, respectively. Therefore, the force resultant on the tool body can be written,
\begin{equation}\label{F*E4}
    ^{G_3}F^*  = {^{G_3}}F - {^{G_3}}U_{T}\, {^T}F,
\end{equation}
where, the net spatial force vector $^{G_3}F^*$ of the tool can be written in the sense of (\ref{RB dynamics}) as,
\begin{equation}
    M_{G_3} \dfrac{\rm d}{\mathrm{d}t} \left( {^{G_3}V} \right) + C_{G_3} \left( {^{G_3}{\omega}}  \right) {^{G_3}V} + G_{G_3} +\, ^{G_3}\Delta_R(t) = {^{G_3} F^*},
    \label{F_G3}
\end{equation}
with \(\Delta_R(t)\) being unknown uncertainty. Considering (\ref{Dtr}), the following hold,

\begin{equation} \label{E4Vr}
    {^{G_3}}V_r = {^{T}}U_{G_3}^T\,{^T}V_r,
\end{equation}
\begin{equation} \label{TFr}
    {^T}F_r = N_c\, f_{ed},
\end{equation}
\begin{equation}
    ^{G_3}F^*_r = Y_{G_3} \hat{\phi}_{G_3} + K_{G_3}\,(^{G_3}V_r - ^{G_3}V)+\,^{G_3}\hat{\Delta}_R,
    \label{F_G3r}
\end{equation}
\begin{equation}
    f_{ed} = M_e \Dot{V}_{sr} + D_e V_s + K_e \mathbf{x}_e,
    \label{f_ed}
\end{equation}
where \(^{G_3}\hat{\Delta}_R\) is in the sense of Lemma \ref{RBFNNs}. According to (\ref{F*E4}), the required spatial force vector, which acts as the control signal for tool body, can be written as,
\begin{equation}\label{E4F_r}
    {^{G_3}}F_r =\, ^{G_3}F^*_r +  {^{G_3}}U_{T}\, {^T}F_r.
\end{equation}
Consequently, the required piston force can be computed as,
\begin{equation}
f_{cr} = \frac{1}{r_{w}}\sigma^T \, {^{ G_{3}} F_r} + \sigma^T {^{ G_{p3}} F_r^*},
\label{wrist piston force req}
\end{equation}
where \(r_{w}\) being gear ratio of helical gears in the actuator and \({^{ G_{p3}} F_r^*}\) being the net required spatial force of piston body in the sense of (\ref{F_G3r}). 
\begin{rmk}
    The first term in (\ref{wrist piston force req}) accounts for the dynamics of the rigid body subsystem, while the second term compensates for the joint mechanism's dynamics and uncertainties, ensuring the fulfillment of the control objectives. The low-level voltage control for generating the required piston force follows the approach outlined in \cite{hejrati2023orchestrated}.
\end{rmk}

\begin{figure}[pos=t]
\centering
\includegraphics[width=.4\textwidth]{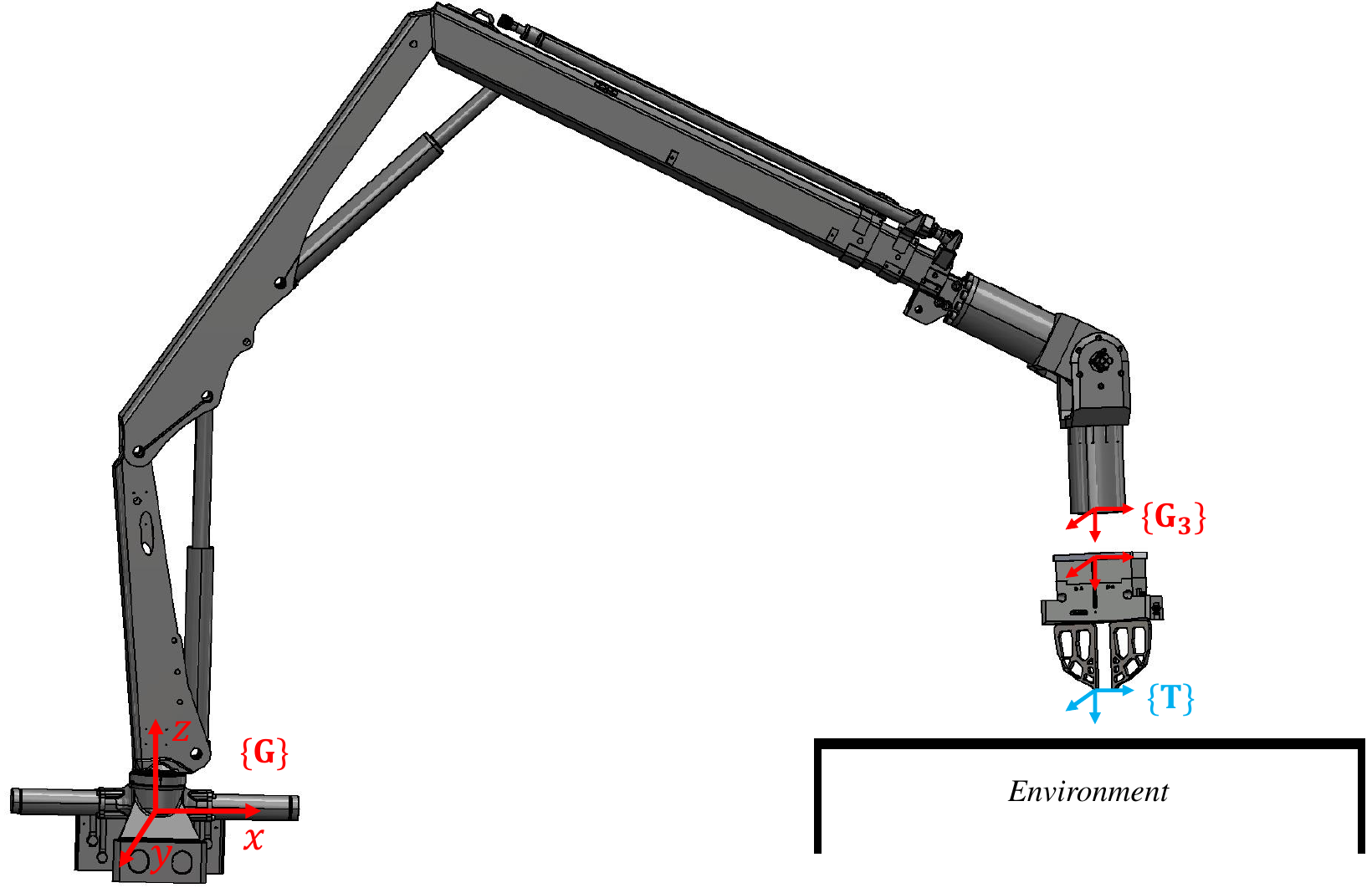}
\caption{ Scheme of 6-DoF HHM in presence of unknown environment }
\label{Hiab}
\end{figure}

\begin{thm}
    The tool of the surrogate with dynamics represented in (\ref{F_G3}) and required net spatial force defined in (\ref{F_G3r}) is virtually SGUUB.
    \label{G3 THM}
\end{thm}
\begin{pf}
    Following the same procedure in \cite[Theorem 2]{hejrati2024impact} for frame $\{G_3\}$ with non-negative accompanying function \(\nu_{G_3}\), the following holds,
    \begin{equation}
        \Dot{\nu}_{G_3} \leq -\,\mu\,\nu_{G_3}+\,\mu_0 + p_{G_3} -p_{T},
        \label{Dnu3}
    \end{equation}
    with \(\mu\) and \(\mu_0\) being positive constants, \(p_{G_3}\) being VPF of driving cutting point, and \(p_{T}\) is the VPF between surrogate end-effector and environment. According to Lemma \ref{lemma SGUUB} and (\ref{Dnu3}), the tool subsystem is virtually SGUUB.
    \end{pf}
When there is no contact with the environment, \(p_{T} = 0\). However, during contact, \(p_{T}\) represents the power flow between the surrogate and the remote environment. The following lemma used to tackle \(p_{T}\) during contact.
    \begin{lem}
        The following holds for \(p_{T}\),
        \begin{equation}
            \lim_{t \to \infty}\int_0^t {p_{T} dt} \geq  -\varrho_{0s},
        \end{equation}
        where \(\varrho_{0s}\) is a positive constant.
        \label{pT}
    \end{lem}
    \begin{pf}
        Appendix \ref{p_T}.
    \end{pf}

\begin{thm}
    Consider the surrogate decomposed into subsystems employing VCP concept. Then, for all \(\mathbf{A} \in \Psi\), where \(\Psi\) encompasses all the rigid body frames attached to the subsystem except $\left\lbrace  G_3 \right\rbrace$, with,
    \begin{equation}
    \begin{split}
	{\nu}_s &= \nu_{G_3} + \sum_{ \mathbf{A} \in S} \nu_{\mathbf{A}} +\sum_{j=1,6}\left( f_{pjr} - f_{pj} \right)^2/(2 \, \beta \, k_{xj})
    \end{split}
	\label{eqn: v function for RB}
    \end{equation}
    all the signals of the closed-loop system are SGUUB. Consequently, the following holds,
    \begin{equation}
    \rho_s \defequal V_{sr}-V_s = V_{sd}-V_s-\mathcal{A}\boldsymbol{F}_s \in \Omega_s,
    \label{rho_s}
\end{equation}
where 
\begin{equation}
    \Omega_s \defequal \lbrace V_{sr}-V_s \in \Re^6|\, ||V_{sr}-V_s||\leq \alpha\sqrt{\frac{2(\Bar{\mu}_0+\,\varrho_{s})}{\lambda_{min}(M_{G_3}) \Bar{\mu}}} \rbrace,
\end{equation}
where \(||^{G_3}U_T^T||\leq \alpha_s\), \(||p_T||\geq\,\varrho_{s}\) with \(\Bar{\mu}\) and \(\Bar{\mu}_0\) being positive constants.
    \label{Thm Hiab}
\end{thm}
\begin{pf}
    The proof follows the same approach as outlined in \cite[Theorem 3]{hejrati2023orchestrated} and Theorem \ref{Thm Exo}, using Lemma \ref{pT}.
\end{pf}

\section{{Immersive Bilateral Teleoperation with Enhanced SoE}}\label{teleoperation}

In this section, the desired velocities of the master and surrogate robots, denoted as \(V_{md}\) and \(V_{sd}\), are designed to enable dissimilar bilateral teleoperation over a communication channel with motion and force scaling. {Furthermore, to enhance the SoE through immersion, the role and implementation of a VR-based visual feedback system are introduced and discussed.}

{We begin by formulating the bilateral teleoperation framework as follows:}

\begin{equation}
\begin{split}
    V_{md} = \kappa_p^{-1}\big( \boldsymbol{V}_s 
    &+ \Lambda \left[\boldsymbol{X}_s - \kappa_p X_m\right] \\
    &- \mathcal{A}\left[\boldsymbol{F}_s + (\kappa_f - \kappa_p)\boldsymbol{F}_m\right] \big),
    \label{Vmd}
\end{split}
\end{equation}
\begin{equation}
\begin{split}
    V_{sd} = \kappa_p \big(\boldsymbol{V}_m + \Lambda\boldsymbol{X}_m\big) 
    - \Lambda X_s - \mathcal{A}\kappa_f\boldsymbol{F}_m,
    \label{Vsd}
\end{split}
\end{equation}
where \(\kappa_p \in \mathcal{R}^{6\times6}\) and \(\kappa_f \in \mathcal{R}^{6\times6}\) are the motion and force scaling matrices, respectively; \(\Lambda \in \mathcal{R}^{6\times6}\) is a positive-definite matrix; and \(X_m\), \(V_m\), \(X_s\), and \(V_s\) represent the pose and velocity of the master robot and the pose and velocity of the surrogate end-effectors, respectively. As can be seen, the designed velocity commands not only incorporate motion and force scaling, but also enable both force reflection and force tracking. It is important to note that, in this study, orientation error is computed based on quaternions.

\subsection{Stability Analysis}
\subsubsection{Stability without delay}
Substituting (\ref{Vmd}) and (\ref{Vsd}) into (\ref{rho_m}) and (\ref{rho_s}), respectively, with \(\Dot{X}_m = V_m\) and \(\Dot{X}_s = V_s\), we can write,
\begin{equation}
\begin{split}
    \rho_s -\kappa_p\rho_m &= \kappa_p \boldsymbol{V}_m -\boldsymbol{V}_s+\Lambda[\kappa_p \boldsymbol{X}_m-\boldsymbol{X}_s]+\kappa_pV_m\\
    &-V_s+\Lambda[\kappa_p {X}_m-{X}_s]\\
    & = \boldsymbol{{e}}+{e},
    \end{split}
    \label{e_sum}
\end{equation}
with,
\begin{equation}
    {e} \defequal \kappa_pV_m-V_s+\Lambda[\kappa_p {X}_m-{X}_s].
\end{equation}
According to results of Theorem \ref{Thm Exo} and \ref{Thm Hiab} along with (\ref{e_sum}), one can obtain,
\begin{equation}
    {e} \in \Omega_T,
\end{equation}
\begin{equation}
    \Omega_T \defequal \lbrace {e} \in \Re^6|\, \|{e}\| \leq \alpha_s \sqrt{\frac{2(\Bar{\mu}_0+\,\varrho_{s})}{\lambda_{min}(M_{G_3}) \Bar{\mu}}}+ \kappa_p \mathcal{D}_m \rbrace,
\end{equation}
leading to,
\begin{equation}
    \rho_v \defequal \kappa_pV_m-V_s \in \Omega_T,
    \label{rho_v}
\end{equation}
\begin{equation}
    \rho_p \defequal \kappa_pX_m-X_s \in \Omega_T.
    \label{rho_p}
\end{equation}
The above equations ensure the boundedness of velocity and position errors during both free-motion and in-contact phases of teleoperation in the presence of unknown uncertainties, human exogenous forces, and environmental forces.

\subsubsection{Stability with delay}
In order to address time delays existing in communication medium between master and surrogate, the expressions in (\ref{Vmd}) and (\ref{Vsd}) can be modified as,
\begin{equation}
    \begin{split}
        V_{md} = \kappa_p^{-1}( e^{-s\mathcal{T}} (\boldsymbol{V}_s &+ \Lambda \boldsymbol{X}_s)-\kappa_p \Lambda X_m\\
        &-\mathcal{A}\left[e^{-s\mathcal{T}} \boldsymbol{F}_s+(\kappa_f-\kappa_p\boldsymbol{F}_m\right] ),
        \label{Vmd_delayed}
\end{split}
\end{equation}
\begin{equation}
    \begin{split}
        V_{sd} = e^{-s\mathcal{T}} \kappa_p (\boldsymbol{V}_m+\Lambda\boldsymbol{X}_m)-\Lambda X_s - e^{-s\mathcal{T}} \mathcal{A}\kappa_f\boldsymbol{F}_m,
    \end{split}
    \label{Vsd_delayed}
\end{equation}
with \(\mathcal{T}\) being one-way arbitrary time delay. Substituting (\ref{Vmd_delayed}) and (\ref{Vsd_delayed}) in (\ref{rho_m}) and (\ref{rho_s}), one can obtain,
\begin{equation}
\begin{split}
        \kappa_p(\boldsymbol{V}_m + \Lambda \boldsymbol{X}_m) + \mathcal{A} \kappa_f\boldsymbol{F}_m = e^{-s\mathcal{T}}[&\boldsymbol{V}_s+\Lambda \boldsymbol{X}_s\\
        &-\mathcal{A}\boldsymbol{F}_s]-\kappa_p \rho_m,
\end{split}
\end{equation}
\begin{equation}
\begin{split}
        V_s + \Lambda X_s + \mathbf{A} \boldsymbol{F}_s = e^{-s\mathcal{T}}[\kappa_p(\boldsymbol{V}_m &+ \Lambda \boldsymbol{X}_m)\\
        &-\mathcal{A} \kappa_f\boldsymbol{F}_m]-\rho_s.
\end{split}
\end{equation}
Then, the stability of the teleoperation system under arbitrary time delay \(\mathcal{T}>0\) can be ensured by satisfying the following sufficient condition,
\begin{equation}
    \left\{
    \begin{aligned}
        &\parallel \mathcal{H}_{1s}(jw)^{-1}\, \mathcal{H}_{2s}(jw)\,\mathcal{H}_{1m}(jw)^{-1}\, \mathcal{H}_{2m}(jw) \parallel_\infty < 1, \\
        &\parallel \mathcal{H}_{1m}(jw)^{-1}\, \mathcal{H}_{2m}(jw)\,\mathcal{H}_{1s}(jw)^{-1}\, \mathcal{H}_{2s}(jw) \parallel_\infty < 1,
    \end{aligned}
    \right.
    \label{Delay condition}
\end{equation}
with,
\begin{equation}
    \mathcal{H}_{1s}(s) \defequal (s\mathbf{I}+\Lambda)(\mathcal{C}^{-1}s+\mathbf{I}_6)+\,s\mathcal{A}\mathbf{Z}_e(s),
\end{equation}
\begin{equation}
    \mathcal{H}_{2s}(s) \defequal (s\mathbf{I}+\Lambda)\mathbf{I}_6 - \,s\mathcal{A}\,\kappa_f\,\kappa_p^{-1}\mathbf{Z}_h(s),
\end{equation}
\begin{equation}
    \mathcal{H}_{1m}(s) \defequal (s\mathbf{I}+\Lambda)(\mathcal{C}^{-1}s+\mathbf{I})+\,s\mathcal{A}\,\kappa_f\,\kappa_p^{-1}\mathbf{Z}_h(s),
\end{equation}
\begin{equation}
    \mathcal{H}_{2m}(s) \defequal (s\mathbf{I}+\Lambda)-\,s\mathcal{A}\mathbf{Z}_e(s),
\end{equation}
where \(\mathbf{Z}_e = F_s(s)/V_s(s)\) and \(\mathbf{Z}_h(s) = (F_m(s)-F_h(s))/V_m(s)\). A simplified model for a single DoF system is derived in \cite{lampinen2021force}.

\subsection{Transparency}
Substituting (\ref{Vmd}) and (\ref{Vsd}) into (\ref{rho_m}) and (\ref{rho_s}), one can obtain,
\begin{equation}
    \begin{split}
    \rho_s +\kappa_p\rho_m =
    & -\mathcal{C}^{-1}\left(\kappa_p \Dot{\boldsymbol{V}}_m+\Dot{\boldsymbol{V}}_s+\Lambda(\kappa_p\boldsymbol{V}_m+\boldsymbol{V}_s)\right)\\
    &-2\mathcal{A}\,(\boldsymbol{F}_s+\kappa_f \boldsymbol{F}_m).
    \end{split}
    \label{e_tra}
\end{equation}
Then, substituting (\ref{rho_v}) and (\ref{rho_p}) in (\ref{e_tra}), one can have,
\begin{equation}
    -\boldsymbol{F}_m = \kappa_f^{-1}\boldsymbol{F}_s +\kappa_f^{-1}\kappa_p \mathcal{A}^{-1} \mathcal{C}^{-1} (s\mathbf{I}+\Lambda) \boldsymbol{V}_m + \dfrac{1}{2}\kappa_f^{-1}\rho_1,
    \label{Fm}
\end{equation}
with
\begin{equation}
    \rho_1 \defequal \mathcal{A}^{-1}\left[\mathcal{C}^{-1}(s\mathbf{I}+\Lambda)(\boldsymbol{\rho}_v)+(\rho_s + \kappa_p \rho_m)\right].
\end{equation}
On the other hand, from Fig. \ref{Block_diagram} following holds,
\begin{equation}
    \boldsymbol{V}_s = -\dfrac{s}{s\mathbf{I}+\Lambda}\mathcal{A}\boldsymbol{F}_s+\kappa_p \boldsymbol{V}_m+\dfrac{s}{s\mathbf{I}+\Lambda}\rho_2 ,
    \label{Vs}
\end{equation}
where,
\begin{equation}
    \rho_2 \defequal -\rho_s + \kappa_f [\rho_m - (V_{md}-V_m)].
\end{equation}

\begin{figure}[pos=t]
\centering
\includegraphics[width=.4\textwidth]{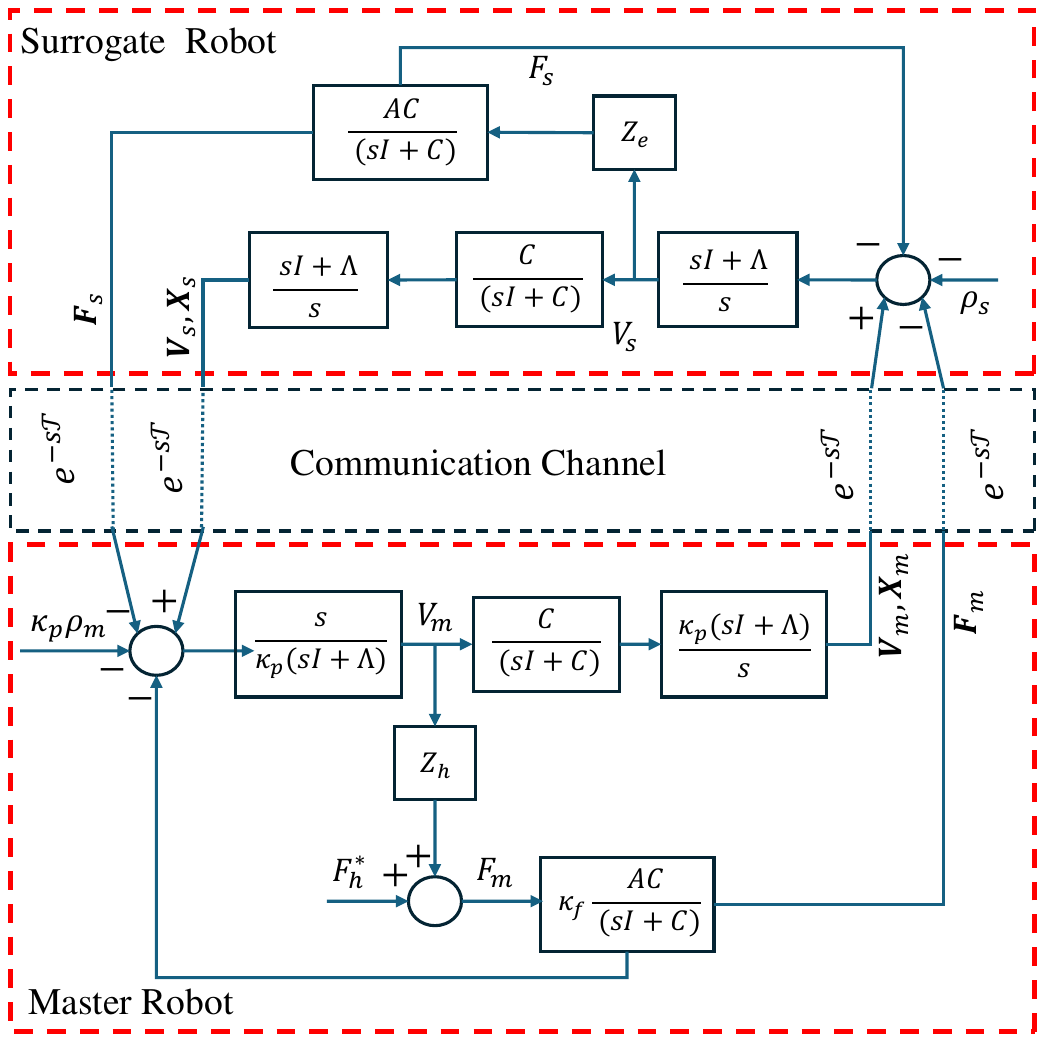}
\caption{ Block diagram representation of dissimilar force-reflected bilateral teleoperation with arbitrary communication delay.} \label{Block_diagram}
\end{figure}

Then, (\ref{Fm}) and (\ref{Vs}) can be written in a compact form as,
\begin{equation*}
\begin{split}
\begin{bmatrix}
-\boldsymbol{F}_m \\ 
-\boldsymbol{V}_s
\end{bmatrix} &= \underbrace{
\begin{bmatrix}
\underbrace{\kappa_f^{-1}\kappa_p \mathcal{A}^{-1} \mathcal{C}^{-1} (s\mathbf{I}+\Lambda)}_{{g_{11}}} & \kappa_f^{-1} \\ 
-\kappa_p & \underbrace{\dfrac{s}{s\mathbf{I}+\Lambda}\mathcal{A}}_{{g_{22}}}
\end{bmatrix}}_{\text{G}}
\begin{bmatrix}
\boldsymbol{V}_m \\ 
\boldsymbol{F}_s
\end{bmatrix}
\end{split}
\end{equation*}
\begin{equation}
\begin{split}
+ \underbrace{\begin{bmatrix}
\dfrac{1}{2}\kappa_f^{-1}\rho_1 \\ 
-\dfrac{s}{s\mathbf{I}+\Lambda}\rho_2
\end{bmatrix}}_{{\rho}}.
\end{split}
\label{G_mat}
\end{equation}
The \(G\) matrix in (\ref{G_mat}) is referred to as the transparency matrix. For high-fidelity bilateral teleoperation with motion/force scaling, the ideal transparency matrix is \cite{hannaford1989design}:
\begin{equation}
    G_{\textbf{ideal}} = \begin{bmatrix}
0 & \kappa_f^{-1} \\ 
-\kappa_p & 0
\end{bmatrix}.
\label{G_ideal}
\end{equation}
A comparison of the ideal transparency matrix in (\ref{G_ideal}) with the actual matrix in (\ref{G_mat}) reveals that ideal transparency can be achieved when \(g_{11} = g_{22} = 0\). However, this criterion is constrained by stability conditions. Consequently, perfect transparency and stability cannot be achieved simultaneously, and only a trade-off between the two can be satisfied. Nevertheless, by appropriately selecting the control gains, a good level of transparency can be achieved.

\begin{rmk}
    Nonzero values of \(g_{11}|_{\boldsymbol{F}_s=0}\) in (\ref{G_mat}) indicate that the operator may experience a sticky sensation while operating the master robot during free motion. Conversely, nonzero values of \(g_{22}|_{\boldsymbol{V}_m=0}\) suggest that during contact between the surrogate and the environment—where the master robot remains stationary—the surrogate will exhibit compliant behavior in response to external forces to ensure stability of the interaction.
\end{rmk}

\begin{rmk}
    The term \(g_{11}\) in (\ref{G_mat}) represents the scaled impedance of the surrogate as perceived by the human arm when there is no contact force (\(\boldsymbol{F}_s = 0\)). From \(g_{11}\), it follows that,
    \begin{equation}
    \begin{split}
        \kappa_f^{-1} \kappa_p \mathcal{A}^{-1} \mathcal{C}^{-1} (s\mathbf{I} + \Lambda)\boldsymbol{V}_m &= \kappa_f^{-1} \left( \mathcal{A}^{-1} \mathcal{C}^{-1} (s\mathbf{I} + \Lambda) \right) \boldsymbol{V}_s \\
        &= \kappa_f^{-1} \mathbf{Z}_s,
    \end{split}
    \end{equation}
    which indicates that during free motion, the force experienced by the human is equivalent to the scaled impedance of the surrogate. This concept is also illustrated in Fig.~\ref{immersive_c}.
\end{rmk}

\begin{figure*}
\centering
\includegraphics[width=0.8\textwidth]{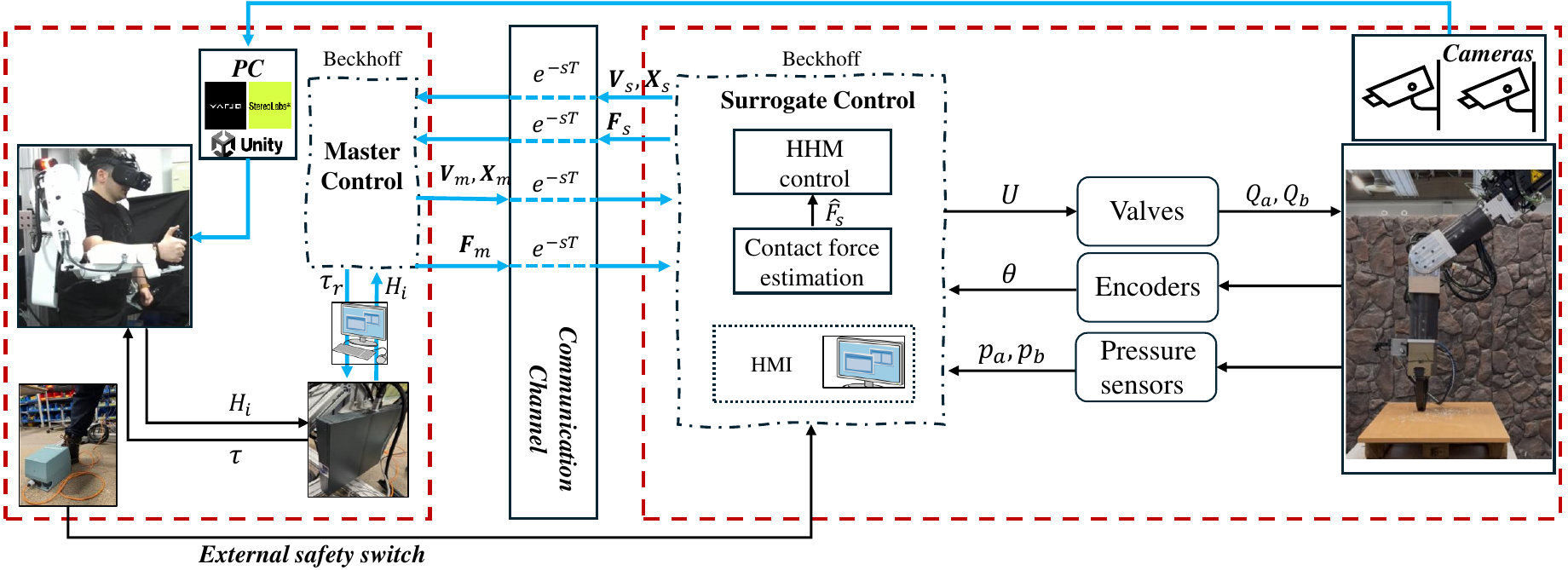}
\caption{ Scheme of experiment setup; The teleoperation setup consists of master robot, master PC, master side Beckhoff, surrogate, surrogate PC, and surrogate side Beckhoff. All the blue lines demonstrate UDP communication, while black lines indicate control signals.} \label{scheme}
\end{figure*}

\subsection{{VR-Based Egocentric Visual Feedback}}

{In the previous sections, we developed a control law to ensure stability and transparency in the bilateral teleoperation of beyond-human-scale robotic manipulators, accounting for time delays, unmodeled uncertainties, and motion/force scaling. These considerations contribute to a higher level of SoE by reinforcing the SoA, which is closely linked to control performance. In this section, we turn our focus to enhancing another core component of the SoE, namely, the SoSL, by integrating immersive hardware into the teleoperation setup, the detail of which is shown in Fig. \mbox{\ref{scheme}}. To the best of our knowledge, this is the first implementation of a fully immersive VR interface with egocentric head-tracking for a beyond-human-scale hydraulic manipulator. The system, illustrated in Fig.~\mbox{\ref{immersive}}, comprises three tightly integrated components: stereo cameras at the remote site, a Unity-based rendering engine, and a VR headset equipped with head and eye tracking.}

{Two ZED Mini stereo cameras operating at 60~fps with a resolution of \mbox{\(2 \times 1280 \times 720\)} are deployed at the remote site: one is fixed near the target location to provide a third-person reference view; the other is an active, pan-tilt stereo camera mounted directly on the robot’s body. This body-mounted camera acts as the egocentric visual source, dynamically synchronized with the operator’s head motion using Unity. As shown in Fig. \mbox{\ref{immersive_b}}, this creates a perspective-aligned visual experience that enhances spatial immersion and realism. To support intuitive interaction, a custom gaze-based interface is implemented within Unity (Fig. \mbox{\ref{immersive_a})}. The operator can switch between available camera views by simply fixating on designated visual icons in the scene (e.g., white gaze circles). This interface leverages the eye-tracking capability of the Varjo XR-3 headset, allowing hands-free control and seamless transition between perspectives.}

{This immersive visual system complements the physical engagement provided by the master interface. While the master robot enables physical interaction with the remote environment, the VR system facilitates cognitive and perceptual immersion, thereby enhancing the operator’s SoSL and overall SoE. The combination of active egocentric view control, multi-camera switching, and gaze-based interaction supports effective situational awareness in highly scaled, complex remote tasks.}


\section{ Experimental Results }\label{results}

This section is dedicated for comprehensive performance evaluation of the proposed method. Fig. \ref{scheme} depicts the detail of the implementation setup. A video demonstrating all the experiments is provided as supplementary material.

\subsection{Experimental Setup}

As can be seen from Fig. \ref{scheme}, our experimental setup is a real-world benchmark designed to rigorously test the proposed control scheme under challenging conditions. The 7-DoF haptic exoskeleton, as the master robot, introduces inherent uncertainties from its complex forearm and wrist mechanisms, including belts, ball screws, and cables. These are compounded by the nonlinear fluid dynamics governing the 6-DoF heavy-duty surrogate. The human operator interfaces with the system through haptic exoskeleton and VR-based active vision, together creating an immersive teleoperation experience representative of real-world industrial applications. This unique configuration not only can challenge the adaptability of the control scheme to mechanical, hydraulic, and network complexities but also evaluates its potential universality and reliability in high-stakes, heavy-duty operations.

The master-side control system is implemented on a Beckhoff CX5130 industrial PC with a 1~ms sampling rate. It communicates with the surrogate-side controller, which runs on a Beckhoff CX2030 unit, also operating at a 1~ms sampling rate. The surrogate robot is actuated through electrohydraulic servo valves with the following specifications: (1) base joint—Bosch 4WRPEH6 valve (12~dm\textsuperscript{3}/min at \(\Delta p = 3.5\)~MPa per notch), (2) lift and tilt joints—Bosch 4WRPEH10 valve (100~dm\textsuperscript{3}/min at \(\Delta p = 3.5\)~MPa per notch), and (3) wrist joints—Bosch 4WRPEH6 valve (40~dm\textsuperscript{3}/min at \(\Delta p = 3.5\)~MPa per notch). Surrogate joint angles are measured using SICK AFS60 absolute encoders (18-bit resolution), and hydraulic pressures are acquired via Unix 5000 pressure transmitters with a maximum operating pressure of 25~MPa.

To ensure the safety of both the human operator and robots during teleoperation, a three-level safety mechanism is implemented. First, in the Human-Machine Interface (HMI), the operator selects the teleoperation mode, only after which the teleoperation is ready to be enabled. Then, the operator has to grab the handle of the exoskeleton, which with its sensing signals, the second phase is enabled and the algorithm is ready for the next. Finally, the pedal shown in Fig. \ref{scheme}, which is an external signal to the control algorithm, enabled the indexing of the master robot. By releasing either pedal or handle of the exoskeleton, the actuators of the exoskeleton are deactivated and the surrogate switches to its local control mode. Such a multi-level safety setup removes the risk of unaware teleoperation connection and potential damage to either robots or the operator. It must be emphasized that, prior to enabling teleoperation, all essential configurations, including control modes and gain adjustments, position and force scaling, and system calibration, are managed through a HMI that has been entirely designed and developed in our laboratory using TwinCAT 3. The HMI is designed to be versatile, enabling communication with other industrial control systems through standardized interfaces. This facilitates integration into broader automation architectures, allowing the system to operate within existing industrial workflows.

\subsection{Scenarios of Experiments}
{The conducted experiments are divided into two main categories: (i) objective evaluation of the proposed method’s performance under varying operational conditions, and (ii) subjective assessment of SoE experienced by the operator.} The first set of experiments evaluates the control performance under different conditions, including various position and force scaling factors, which directly affect the operational velocity and contact force ranges, as well as under different levels of communication delay. To objectively assess the performance of the control scheme, we define the following evaluation metrics:

\begin{itemize}
    \item \textbf{Task Completion Time} (\(\mathscr{T}\)): The total time required for the operator to execute the task.
    \item \textbf{Maximum Tracking Error} {of the pose norm where \mbox{\(\mathscr{N} = [\mathscr{N}_p, \mathscr{N}_o] \in \mathcal{R}^2\)} with \mbox{\(\mathscr{N}_p = ||\kappa_p\,X_m|| - ||X_s||\in \mathcal{R}\)} and \mbox{\(\mathscr{N}_o = ||Er(O)||\in \mathcal{R}\)} where \mbox{\(Er(O)\)} is the quaternion-based orientation error:}
\begin{equation}
        \mathscr{N}_{\text{max}} = \max_{t \in [0, \mathscr{T}]} |\mathscr{N}(t)|.
\end{equation}
    \item \textbf{Root Mean Square Error (RMSE) of the Displacement Norm} (\(\mathscr{N}_{rms}\)):
\begin{equation}
    \mathscr{N}_{rms} = RMS(\mathscr{N}(t))= \sqrt{\frac{1}{N}\sum_{i=1}^N \mathscr{N}_i^2}.
\end{equation}
    \item \textbf{Integral Absolute Error (IAE) of the Displacement Norm} (\(\mathscr{N}_I\)):
\begin{equation}
    \mathscr{N}_I = \int_0^\mathscr{T} |\mathscr{N}(t)| \ dt.
\end{equation}
    This metric captures the total accumulated deviation in position error over the task duration.
    \item \textbf{Mean Absolute Error (MAE) of Velocity Tracking} (\(\overline{\mathrm{V}}\)), defined as the difference between the norm of the master and surrogate velocities:
\begin{equation}
        \overline{\mathrm{V}}  = \frac{1}{N}\sum_{i=1}^N |\left( ||\kappa_p V_m||_i - ||V_s||_i \right)|.
\end{equation}
    \item \textbf{Position-Velocity Normalizing Index} (\(\rho\)), which evaluates the relationship between the maximum position error and the maximum velocity:
\begin{equation}
        \rho = \frac{\mathscr{N}_{\text{max}}}{||V_s||_{\text{max}}}.
\end{equation}
    This index helps assess how well the surrogate can maintain accurate tracking under various velocity conditions.
    \item \textbf{Maximum and RMS of Force Tracking Error} (\(\mathscr{F}_{\mathrm{max}}, \mathscr{F}_{\mathrm{rms}}\)), evaluating the error between the master-side commanded force and the surrogate-side estimated applied force in the direction of interaction where we define {\mbox{\(\mathscr{F} = \kappa_f F_m(t) - F_s(t)\)}}:
\begin{equation}
    \mathscr{F}_{\mathrm{max}} = \max_{t \in [0, \mathscr{T}]} |\mathscr{F}|, \quad
\mathscr{F}_{\mathrm{rms}} = RMS(\mathscr{\mathscr{F}}).
\end{equation}
\end{itemize}

{These metrics enable fair benchmarking of the proposed method and provide a reference point for future studies. In the second set of experiments, a user study is conducted to assess the SoE enabled by the proposed teleoperation platform. To compare the effect of immersive feedback, participants performed the same teleoperation task under two visual conditions: (i) using the immersive VR-based interface and (ii) using only stereo camera-based monitor feedback.}

In all experiments, the filter constant was set to \(\mathcal{C} = 35\). For experiments without time delay, the force and position control gains were set to \(\mathcal{A} = 80 \times 10^{-5}\) and \(\Lambda = 12\), respectively. In contrast, for experiments involving communication delay, the gains were adjusted to \(\mathcal{A} = 50 \times 10^{-5}\) and \(\Lambda = 3\) to satisfy the stability condition defined in~(\ref{Delay condition}), thereby achieving an optimal trade-off between maximum tolerable fixed delay and acceptable control accuracy. For the user study experiments, the control gains were set to \(\mathcal{A} = 80 \times 10^{-5}\) and \(\Lambda = 4\) to ensure smooth teleoperation and allow the surrogate to tolerate fast and high-jerk commands from participants, who are non-expert teleoperators. The local control gains for both the master and surrogate systems were configured identically to those reported in~\cite{hejrati2023physical} and~\cite{hejrati2023orchestrated}.

\begin{figure}[pos=t]
\centering
\includegraphics[width=0.45\textwidth]{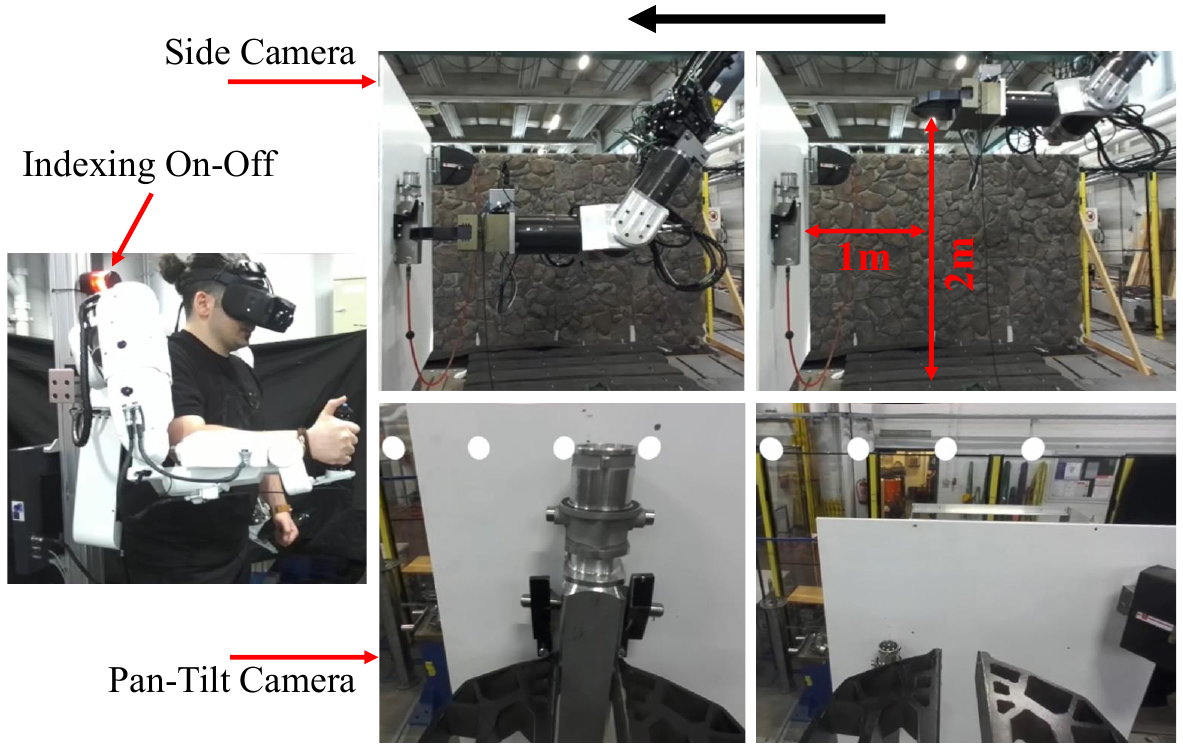}
\caption{ {Sequential steps of the RT experiment, shown from right to left. The surrogate, controlled by a human operator, approaches and grasps the target object.}} 
\label{FM tests}
\end{figure}
\begin{figure*}[b]
    \centering
    \subfloat[{\scriptsize \(\kappa_p = 1\)}]{
    \includegraphics[width=0.31\textwidth]{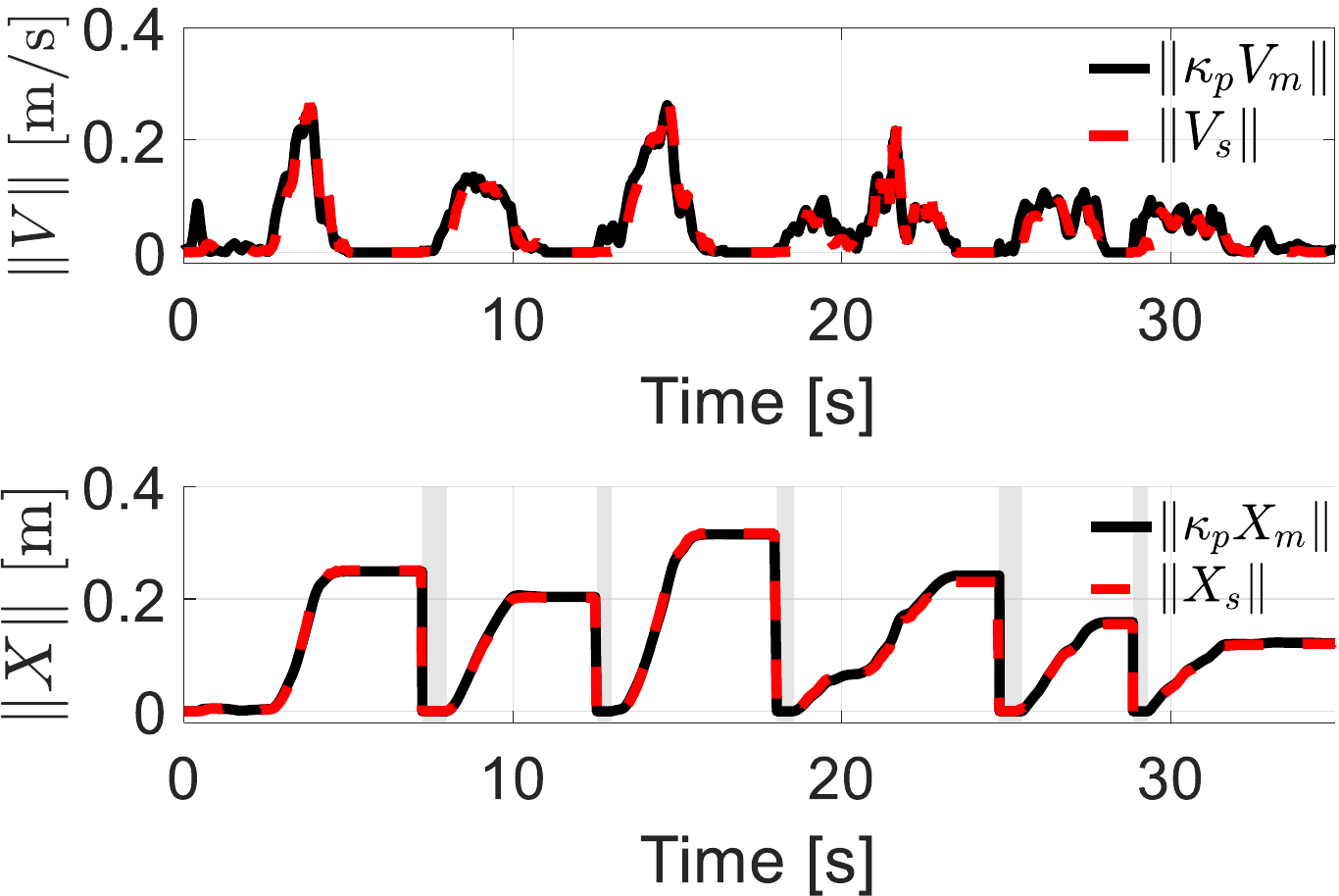}
    \label{FMa}
    }
    \subfloat[\scriptsize \(\kappa_p =7\)]{
        \includegraphics[width=0.3\textwidth]{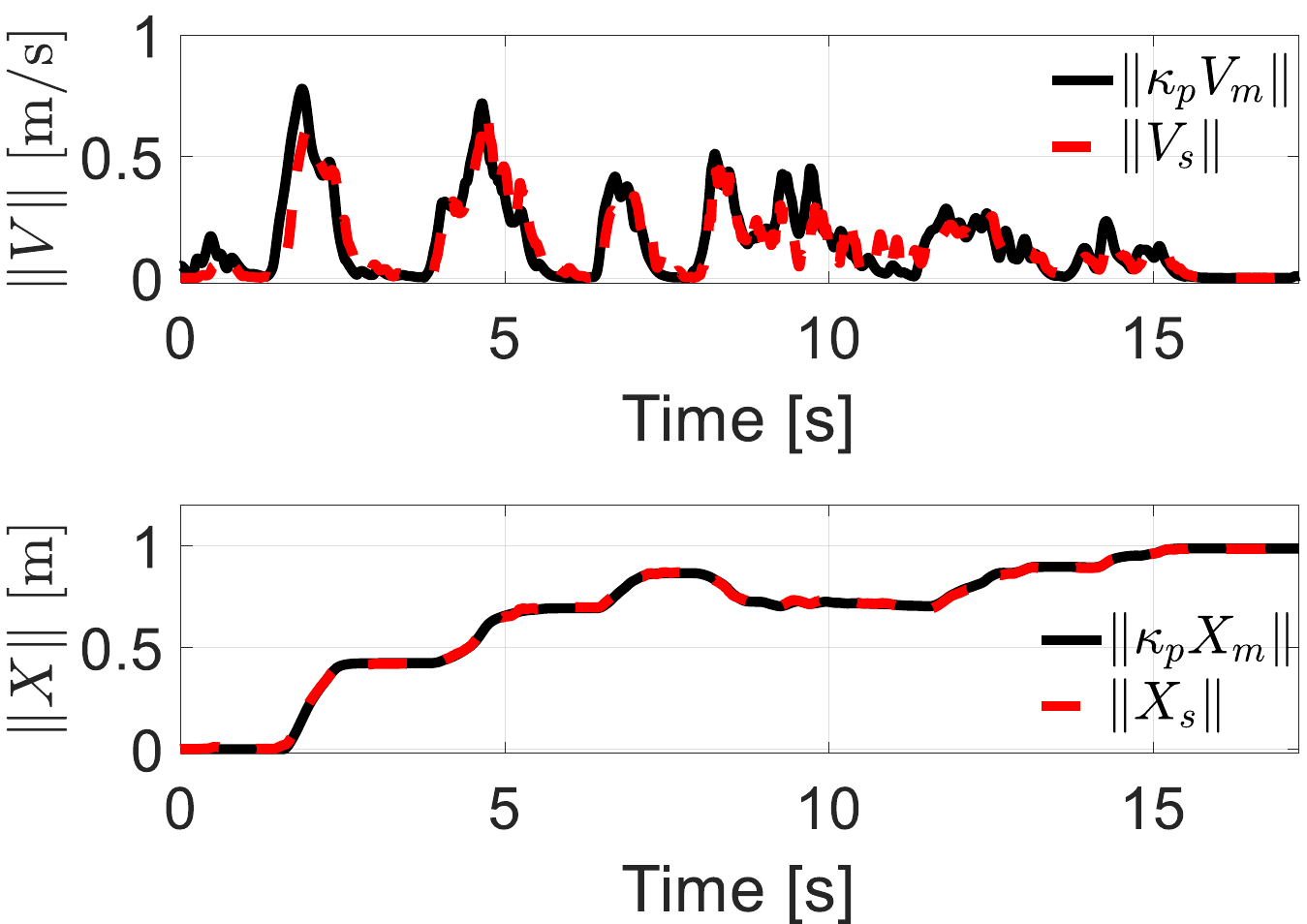}
        \label{FMb}
    }
    \subfloat[\scriptsize \(\kappa_p = 13\)]{
        \includegraphics[width=0.3\textwidth]{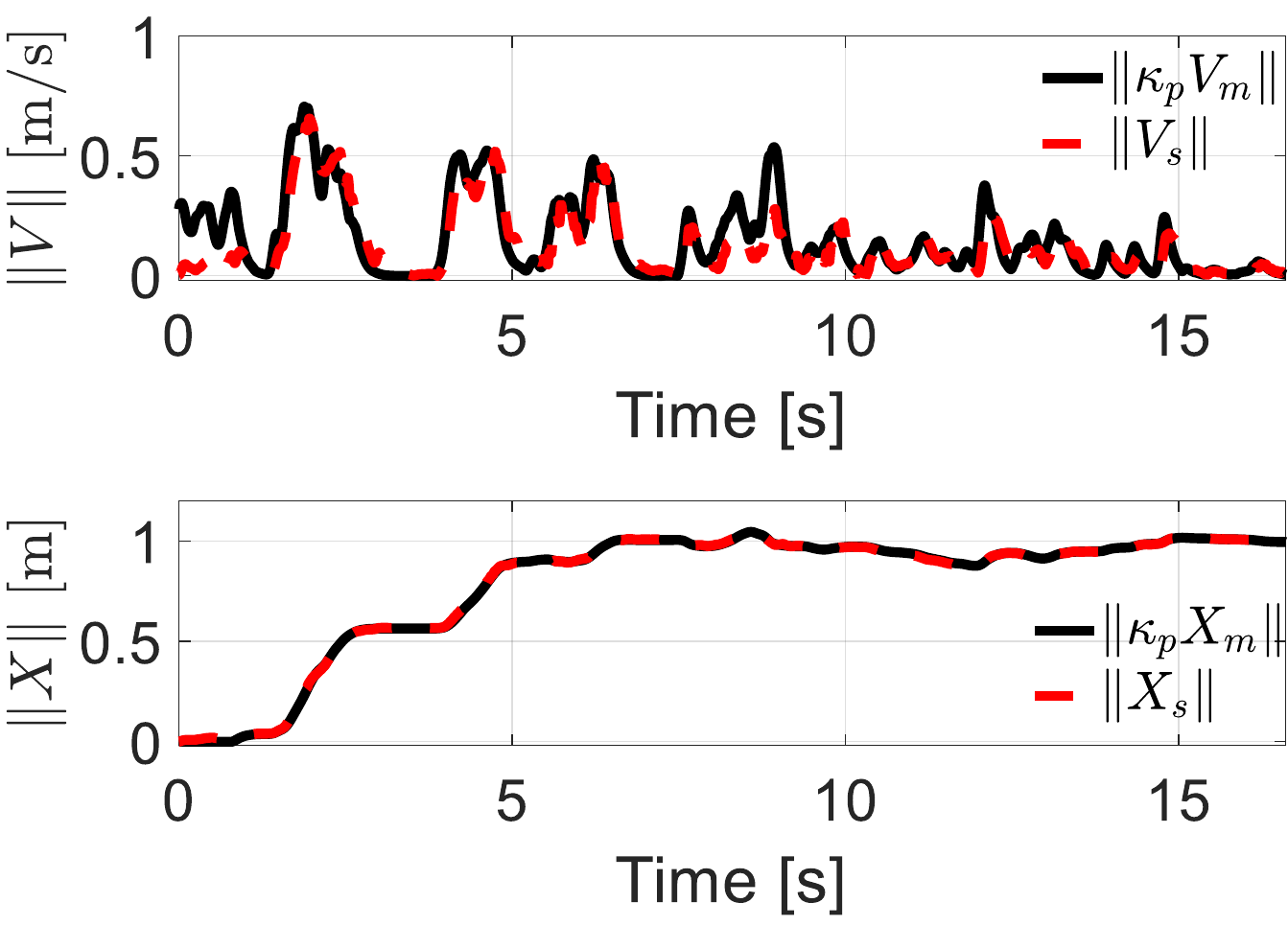}
        \label{FMc}
    }
    \caption{Experimental results of the free-motion task under different motion scaling factors. The shaded area indicates the indexing periods.}
    \label{Motion_track}
\end{figure*}
\subsection{ Free Motion Performance Evaluation}

{This section comprises three sets of experiments. The first, referred to as the \mbox{\textit{reaching task}} (RT), is conducted to evaluate the performance of the proposed method under three different position scaling factors: low (\mbox{\(\kappa_p = 1\)}), medium (\mbox{\(\kappa_p = 7\)}), and high (\mbox{\(\kappa_p = 13\)}). The second set, called the \mbox{\textit{complex pick-and-place task}} (CPT), is designed to assess the method's effectiveness in performing more intricate teleoperation tasks using a beyond-human-scale robotic manipulator, where subtlety and precision are critical. This task is conducted under \mbox{\(\kappa_p = 2\)}. The final experiment, the \mbox{\textit{delay task}} (DT), investigates the controller's performance under two types of communication delay: a fixed delay of \mbox{\(\mathcal{T}_f = 150~\text{ms}\)} and a time-varying delay within the range \mbox{\(0 < \mathcal{T}_v < 150~\text{ms}\)}. The DT is applied to the RT scenario with a scaling factor of \mbox{\(\kappa_p = 3\)}. As outlined above, these experiments are designed not only to assess the controller across different task complexities, but also to evaluate its robustness under various motion scaling conditions and communication delays.}

\begin{figure*}[b]
    \centering
    \includegraphics[width=0.9\textwidth]{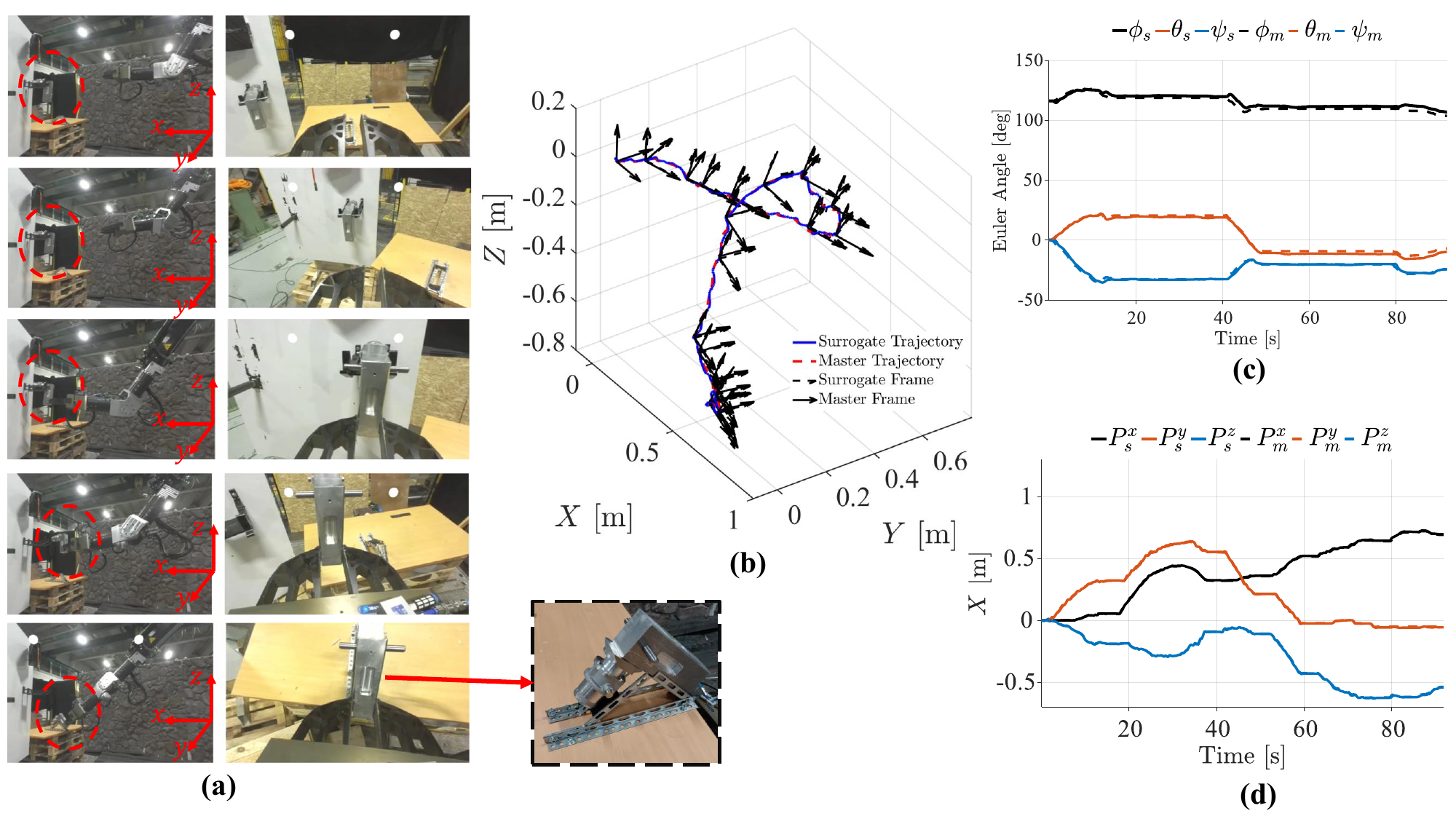}
    \caption{{Results of CPT experiment. (a) The sequence of CPT experiment, (b) 3D path traversed by master and surrogate robots during the task, (d) Euler angle representation of orientation tracking, (b) position tracking.}
    \label{CPT}}
\end{figure*}

\subsubsection{{RT experiments}}
The experimental setup of RT experiments is depicted in Fig. \ref{FM tests} with results presented in Fig.~\ref{Motion_track}. Figs.~\ref{FMa}--\ref{FMc} plot the norm of the end-effector position and velocity for each scale. Across all cases, the proposed controller demonstrates stable tracking and accurate translation of the operator’s commands to the surrogate. Under \(\kappa_p = 1\), the operator required multiple indexing motions to reach the goal, primarily due to the mismatch between the limited workspace of the human arm and the extended reach of the surrogate. In contrast, with higher scaling factors, the operator was able to move the surrogate faster using smaller teleoperation commands. However, this increased mobility came at a perceptual cost. As the scale increases from 7 to 13, operator experienced a temporary mismatch between their proprioceptive sense of motion and the observed robot behavior through VR, which resulted in a longer time to think about how to accomplish the task. This phenomenon is evident in the only marginal improvement in execution time between \(\kappa_p = 7\) and \(\kappa_p = 13\), despite the latter offering greater kinematic amplification. We posit that the observed discrepancy in perceptual adaptation stems from a form of sensorimotor mismatch: while the operator consciously understands the presence of motion scaling, their sensorimotor system temporarily resists reconciling slow proprioceptive input with fast visual feedback from the robot. This mismatch, although initially disorienting, tends to decrease over time with continued experience, and may be further alleviated by introducing longer adaptation phases.

\subsubsection{{CPT Experiment}}
{Fig.~\mbox{\ref{CPT}} illustrates the experimental setup and results of the CPT operation. This task requires the operator to perform precise and coordinated end-effector motions involving significant rotation and translation to align with and grasp the target object. Once grasped, the object must be carefully transported and placed at a designated location. Fig.~\mbox{\ref{CPT}}(a) shows the sequence of task execution, where the object of interest is marked with a red circle for clarity. Fig.~\mbox{\ref{CPT}}(b) presents the 3D trajectories of the master and surrogate end-effectors during the CPT task, visualized with position traces and overlaid orientation frames. The plotted paths show the spatial evolution of both manipulators throughout the task, while the attached frames represent the local coordinate axes of the end-effectors at sampled points. These frames provide visual insight into the rotational alignment between the master and surrogate robots. Figs.~\mbox{\ref{CPT}}(c) and~\mbox{\ref{CPT}}(d) provide detailed tracking of orientation and position, respectively, for both the master and surrogate robots. For clearer representation of orientation, Euler angles are used following the XYZ convention, corresponding to \mbox{\(\phi\)}, \mbox{\(\theta\)}, and \mbox{\(\psi\)}, respectively. As demonstrated, the proposed control method successfully tracks the operator’s commands throughout the operation, leading to accurate and robust task execution. The ability to accomplish such a complex manipulation task using a beyond-human-scale hydraulic manipulator underscores the effectiveness of the control architecture and immersive interface. Notably, this level of performance relies on a high SoA and immersive, real-time visual feedback, both of which are enabled by the proposed teleoperation framework.}

\subsubsection{{DT Experiment}}
{Fig.~\mbox{\ref{delay}} presents the results of the delay tolerance experiments, in which system stability was tested under both fixed and variable communication delays. Fig.~\mbox{\ref{delay}}(a) demonstrates the result with fixed-time delay of 150 ms, while Fig.~\mbox{\ref{delay}}(b) illustrates the result of experiments with variable time delay, which is generated using a uniformly distributed random process. In both delay scenarios, the system maintained stability and completed the task with acceptable accuracy, demonstrating the robustness of the control strategy to unpredictable latency.}

\begin{figure}[pos=t]
    \centering
    \includegraphics[width=0.35\textwidth]{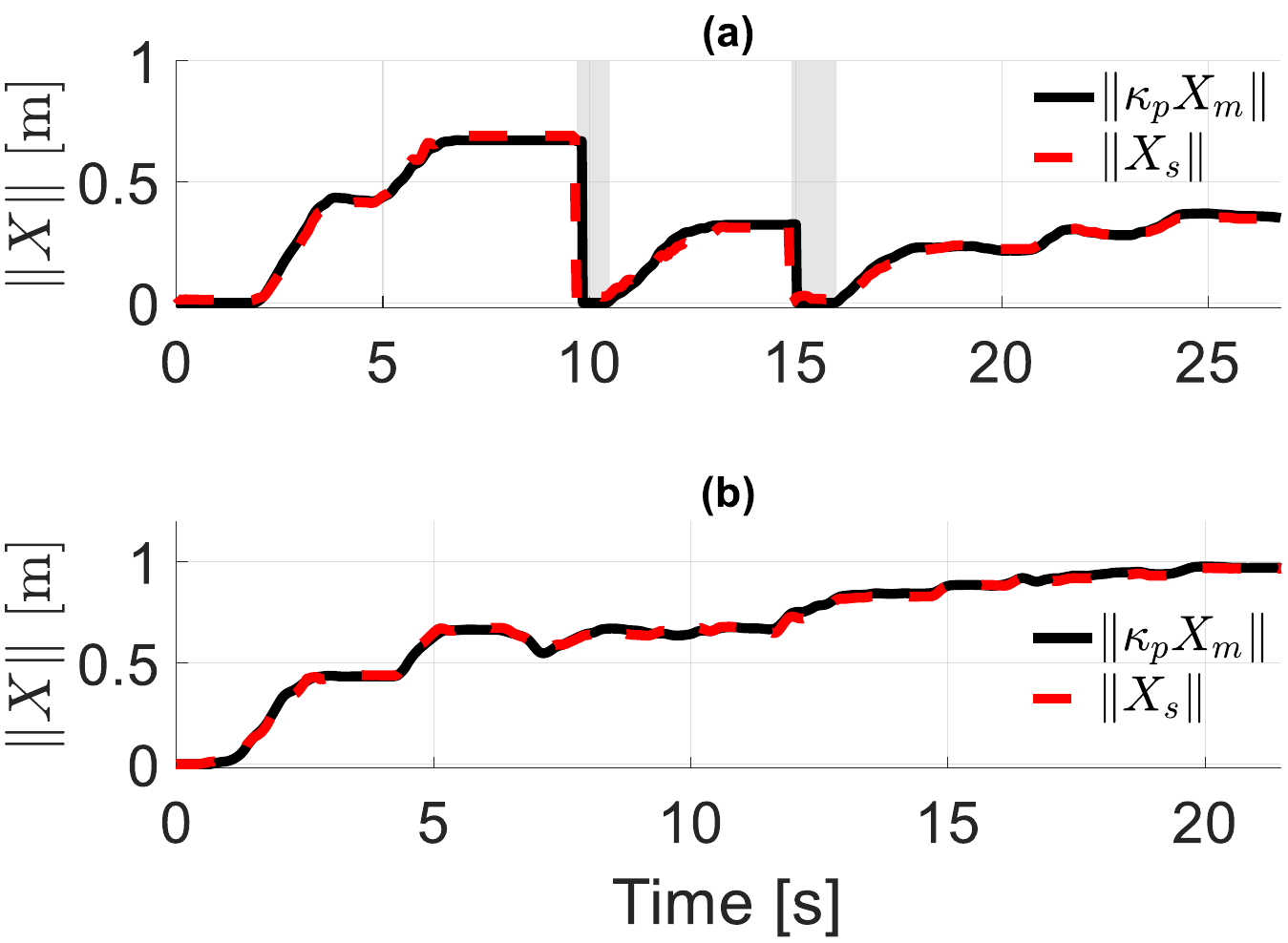}
    \caption{Performance evaluation of the teleoperation system under communication delay with \(\kappa_p =3\). (a) Fixed communication delay \(\mathcal{T}_f = 150~\mathrm{ms}\). (b) Time-varying delay \(0 < \mathcal{T}_v < 150~\mathrm{ms}\).}
    \label{delay}
\end{figure}

\subsubsection{{Quantitative Analysis}}
Table~\ref{tab:performance_metrics} quantifies these observations. For the RT experiments, as expected, task execution time \(\mathscr{T}\) decreases with increasing scale due to faster end-effector velocities in RT experiments. However, the position tracking error increases accordingly: maximum error rises from 1.1~cm to 2.9~cm, and RMS error from 0.37~cm to 0.57~cm, as \(\kappa_p\) increases from 1 to 13. Despite this, the velocity-normalized error index \(\rho\) remains relatively consistent across all experiments, indicating that the tracking accuracy scales proportionally with motion speed, and the controller maintains a well-balanced performance. Furthermore, the controller maintained excellent velocity and orientation tracking performance, even under the high accelerations and velocities induced by large motion scaling factors in such a heavy-duty system.

{In the CPT experiment, the total execution time increased compared to other tasks, due to the complexity of the required motion and the higher number of operator decision-making steps. This task was performed after two initial trials by the operator, which led to a reduction in execution time by approximately one third, indicating rapid operator adaptation and potential platform usability. Despite the increased task complexity and large end-effector motions, the proposed control method maintained accurate and stable behavior throughout the experiment. Quantitative metrics confirm that the maximum position and orientation errors remained within 1.5 cm and 2.5 degrees, respectively. These results suggest that the system provided sufficient SoA for the operator, enabling focus on task execution rather than control compensation.}

{For the DT experiments, the performance metrics reported in Table~\mbox{\ref{tab:performance_metrics}} show that although control accuracy degrades under delay, the controller preserves responsiveness. Notably, fixed-time delay imposes greater performance challenges than time-varying delay, as reflected in higher error magnitudes and longer execution times. Nevertheless, the system remained stable throughout operation under both delay conditions. This characteristic is critical for scenarios in which the robot must be safely guided to a secure state, even under degraded communication conditions.}

\begin{table*}[t]
\centering
\caption{Performance metrics for different motion scaling factors.}
\label{tab:performance_metrics}
\begin{tabular}{ccccccccc}
\hline
\textbf{Experiments} & \(\mathscr{T}\) [s] & \(\mathscr{N}_{p,\text{max}}\) [m] & \(\mathscr{N}_{p,rms}\) [m] & \(\mathscr{N}_{p,I}\) [m$\cdot$s] & \(\overline{\mathrm{V}}\) [m/s] & \(\rho\) & \(\mathscr{N}_{o,\text{max}}\) [deg] & \(\mathscr{N}_{o,rms}\) [deg] \\
\hline
\(\kappa_p = 1\)   & $ 35 $ & $ 0.011 $ & $ 0.0037 $ & $ 0.093 $ & $ 0.011 $ & $ 0.043 $ & $ 0.94 $ & $ 0.34 $ \\ 
\hline
\(\kappa_p = 7\)   & $ 17.25 $ & $ 0.020 $ & $ 0.0052 $ & $ 0.068 $ & $ 0.04 $ & $ 0.03 $ & $ 0.820 $ & $ 0.43 $ \\ 
\hline
\(\kappa_p = 13\)  & $ 16.6 $ & $ 0.029 $ & $ 0.0057 $ & $ 0.064 $ & $ 0.056 $ & $ 0.044 $ & $ 0.76 $ & $ 0.43 $ \\ 
\hline
\multicolumn{9}{l}{{\textit{Experiment below is for CPT:}}} \\
\hline
{\mbox{\(\kappa_p = 2\)}} & {$ 91.7 $} & {$ 0.015 $} & {$ 0.0056 $} & {$ 0.45 $} & {$ 0.024 $} & {$ 0.065 $} & {$ 2.5$} & {$ 0.81 $} \\ 
\hline
\multicolumn{9}{l}{\textit{Experiments below are conducted with delay and \(\kappa_p = 3\):}} \\
\hline
Fixed delay \(\mathcal{T}_f = 150~\mathrm{ms}\) & $ 26.76 $ & $ 0.0430 $ & $ 0.0165 $ & $ 0.37 $ & $ 0.03 $ & $ 0.097 $ & $ 1.32 $ & $ 0.61 $ \\ 
\hline
Time-varying delay \(0 < \mathcal{T}_v < 150~\mathrm{ms}\) & $ 21.47 $ & $ 0.035 $ & $ 0.011 $ & $ 0.2 $ & $ 0.038 $ & $ 0.047 $ & $ 1.41 $ & $ 0.63 $ \\ 
\hline
\end{tabular}
\end{table*}

\begin{figure}[pos=h]
\centering
\includegraphics[width=.3\textwidth]{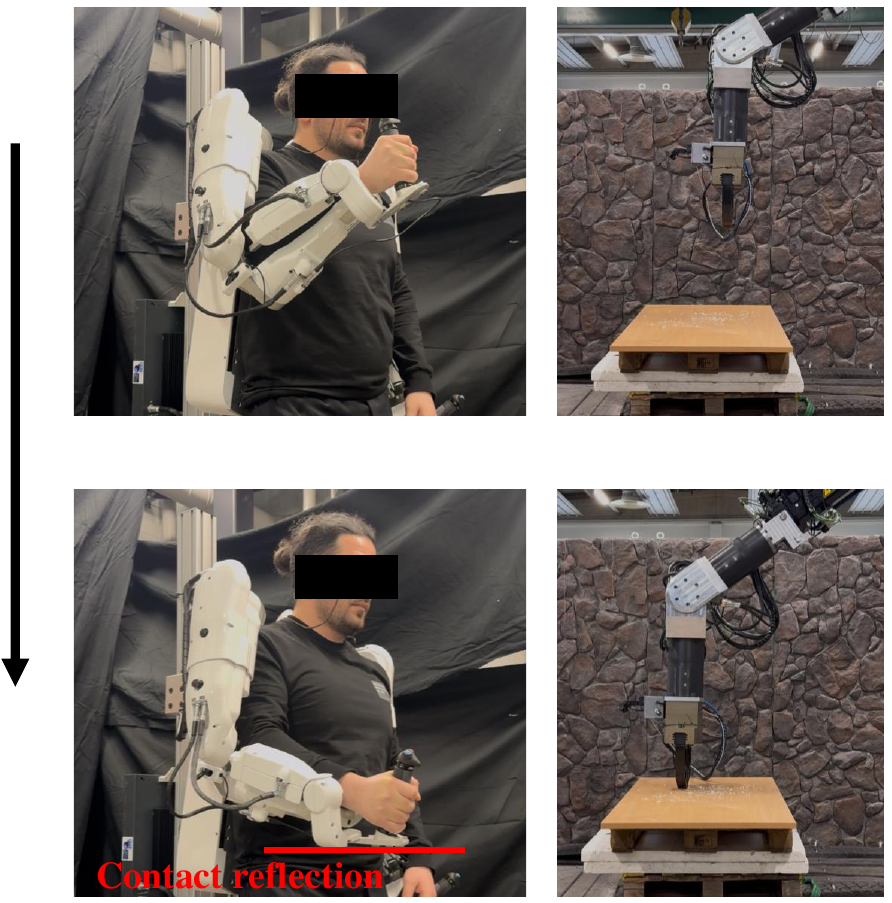}
\caption{ Contact experiment sequences. The operator does not see the contact, and the contact is felt only by the reflection on the arm.} 
\label{Contact test}
\end{figure}

\subsection{ Contact-rich Performance Evaluation}
In this section, the force-reflection and force-tracking capabilities of the teleoperation system are evaluated. To assess these functionalities, the operator was instructed to drive the surrogate, establish contact with the remote environment, and then apply interaction forces upon sensing contact. {To better evaluate the force-reflection, we have conducted three sets of experiments. The first set of experiment, shown in Fig. \mbox{\ref{Contact test}}, is designed to evaluate the performance of the algorithm in different force scaling of \mbox{\(\kappa_f = 500\)}, \mbox{\(800\)}, and \mbox{\(1000\)}. The second set of experiments are performed to assess the performance in presence of communication delay. The last set of experiments are performed to evaluate the multi-axes contact with the environment.} These experiments were conducted with a motion scaling factor of \(\kappa_p = 2\).

\begin{figure*}[h]
    \centering
    \subfloat[]{
        \includegraphics[width=0.28\textwidth]{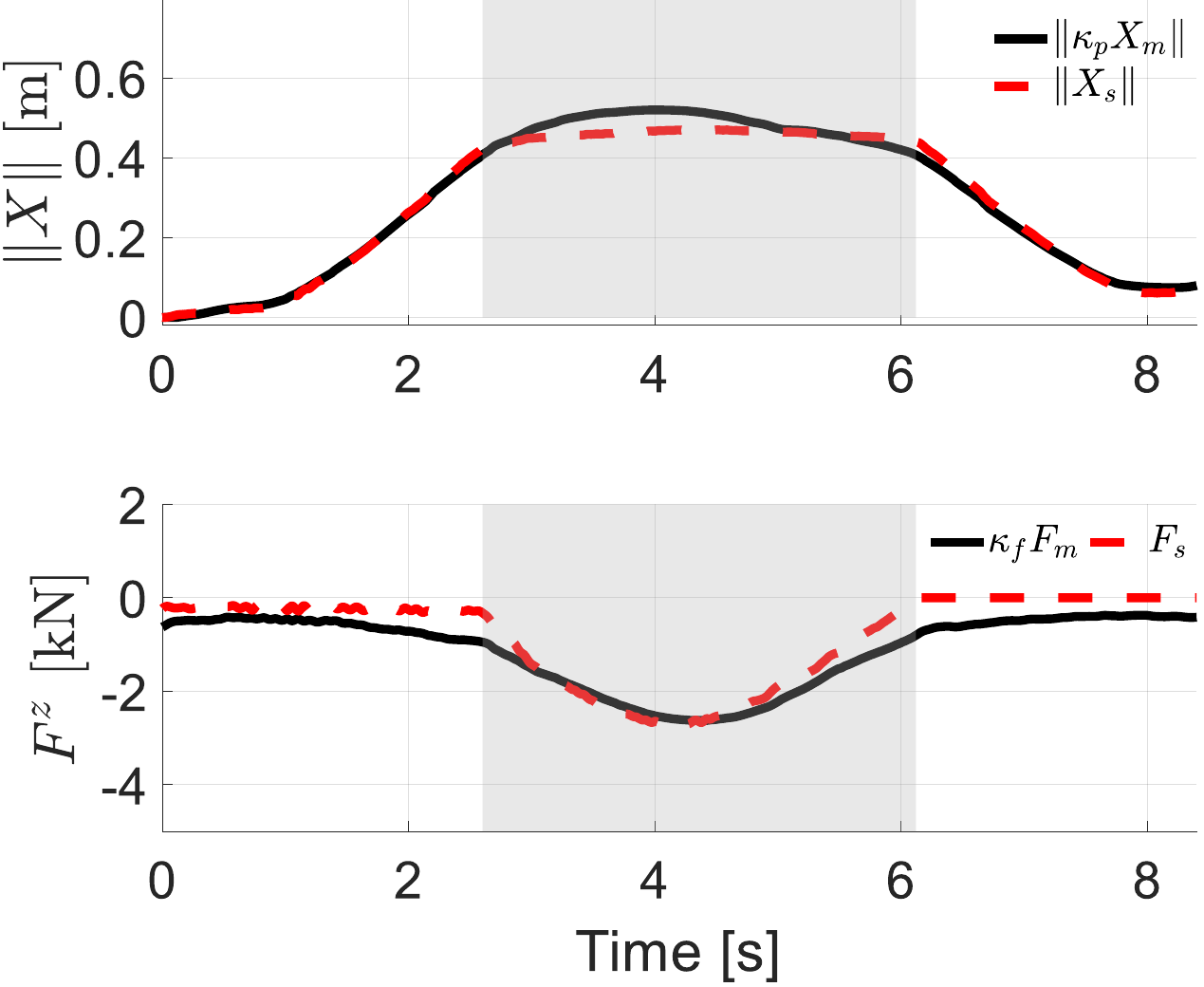}
    }
    \subfloat[]{
        \includegraphics[width=0.28\textwidth]{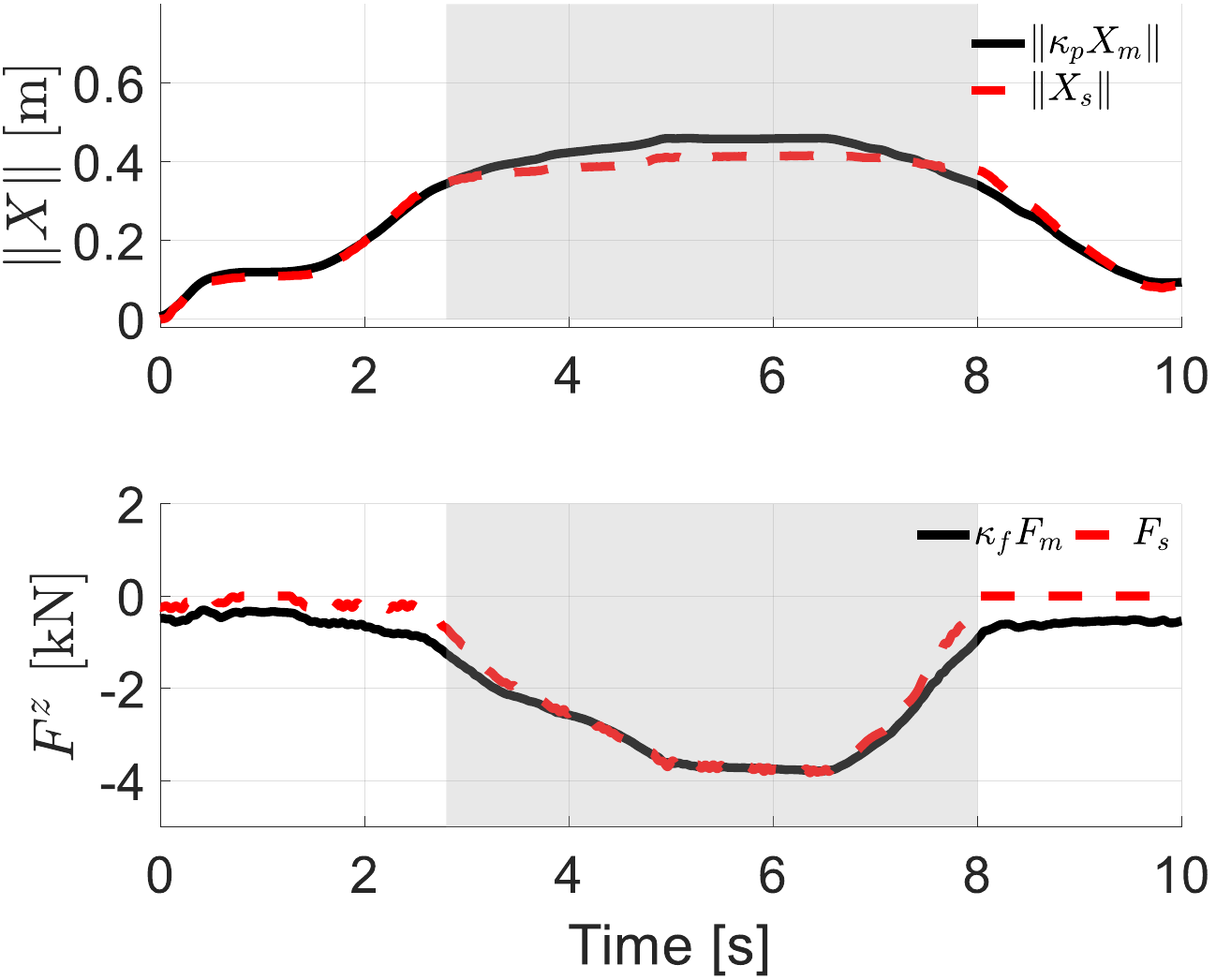}
    }
    \subfloat[]{
        \includegraphics[width=0.28\textwidth]{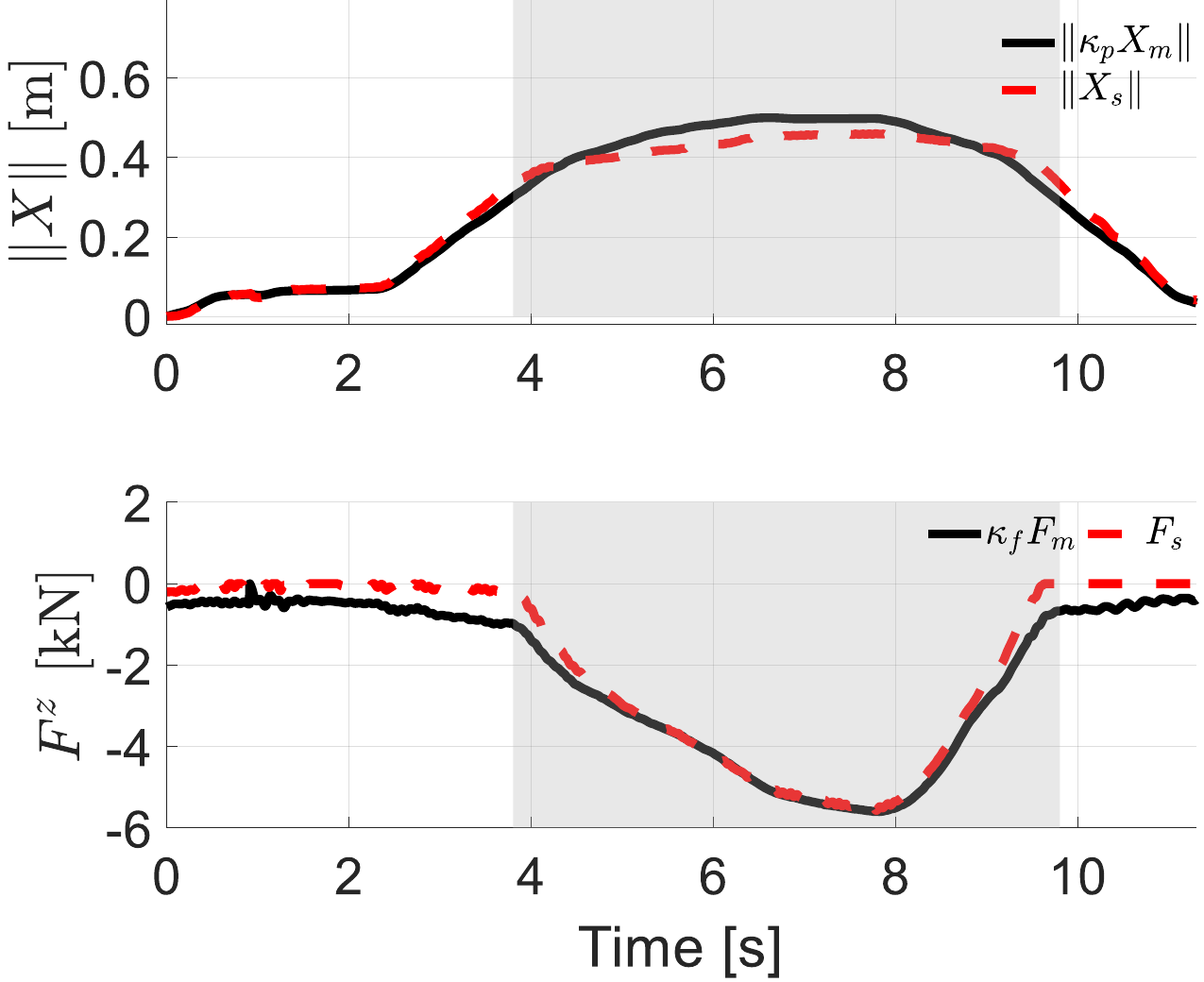}
    }
    \caption{Experimental evaluation of force-reflection and force-tracking performance in teleoperation with \(\kappa_p = 2\). (a) Force and position tracking for \(\kappa_f = 500\); (b) for \(\kappa_f = 800\); (c) for \(\kappa_f = 1000\).}
    \label{contact_task}
\end{figure*}

\begin{figure}[pos=h]
    \centering
    \includegraphics[width=.49\textwidth]{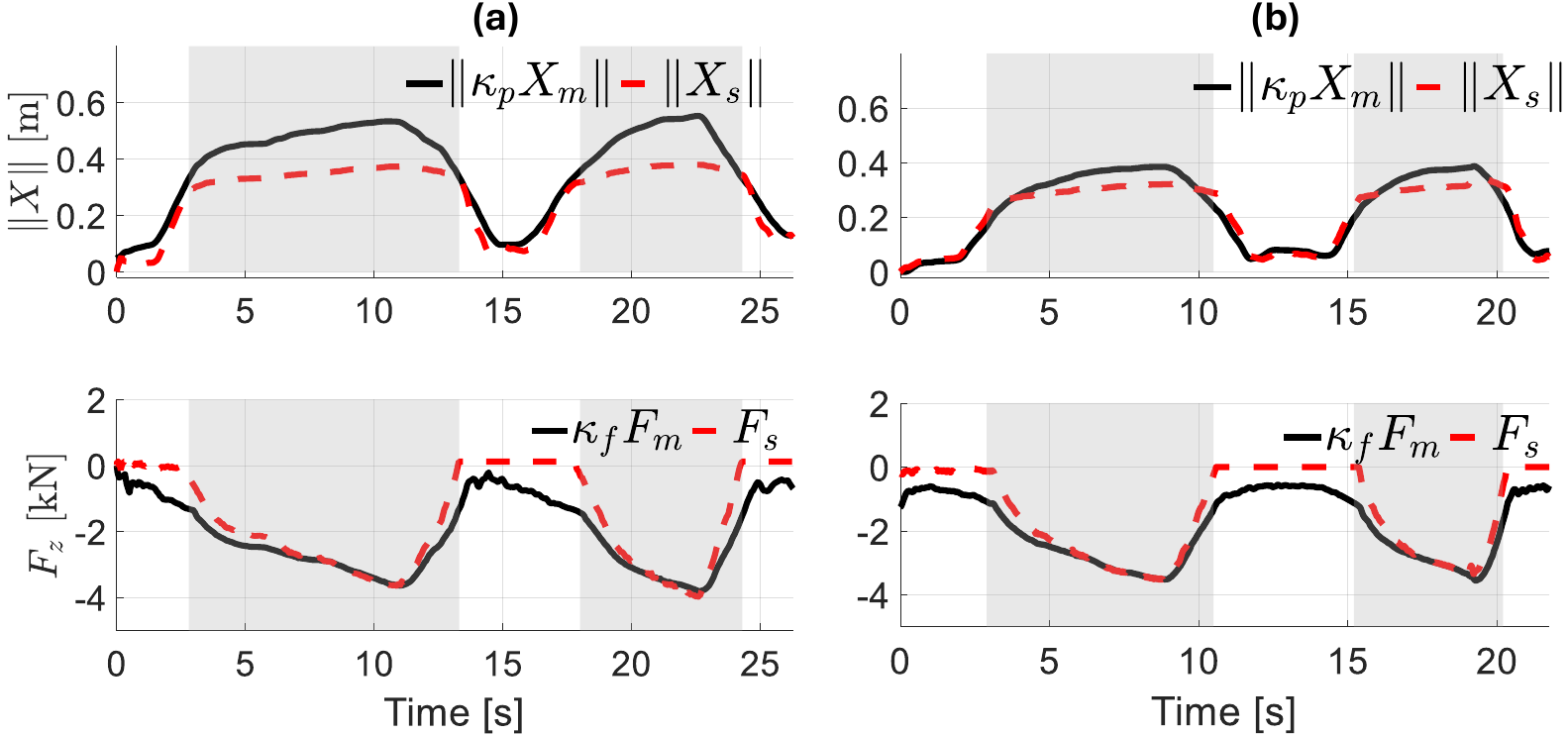}
    \caption{Contact experiments under communication delay with \(\kappa_f = 800\) and \(\kappa_p = 2\). (a) Position and force tracking with fixed delay \(\mathcal{T}_{f} = 150~\mathrm{ms}\); (b) position and force tracking with time-varying delay \(0 < \mathcal{T}_{v} < 150~\mathrm{ms}\).}
    \label{contact_delay}
\end{figure}

\begin{figure}[pos=h]
\centering
\includegraphics[width=.4\textwidth]{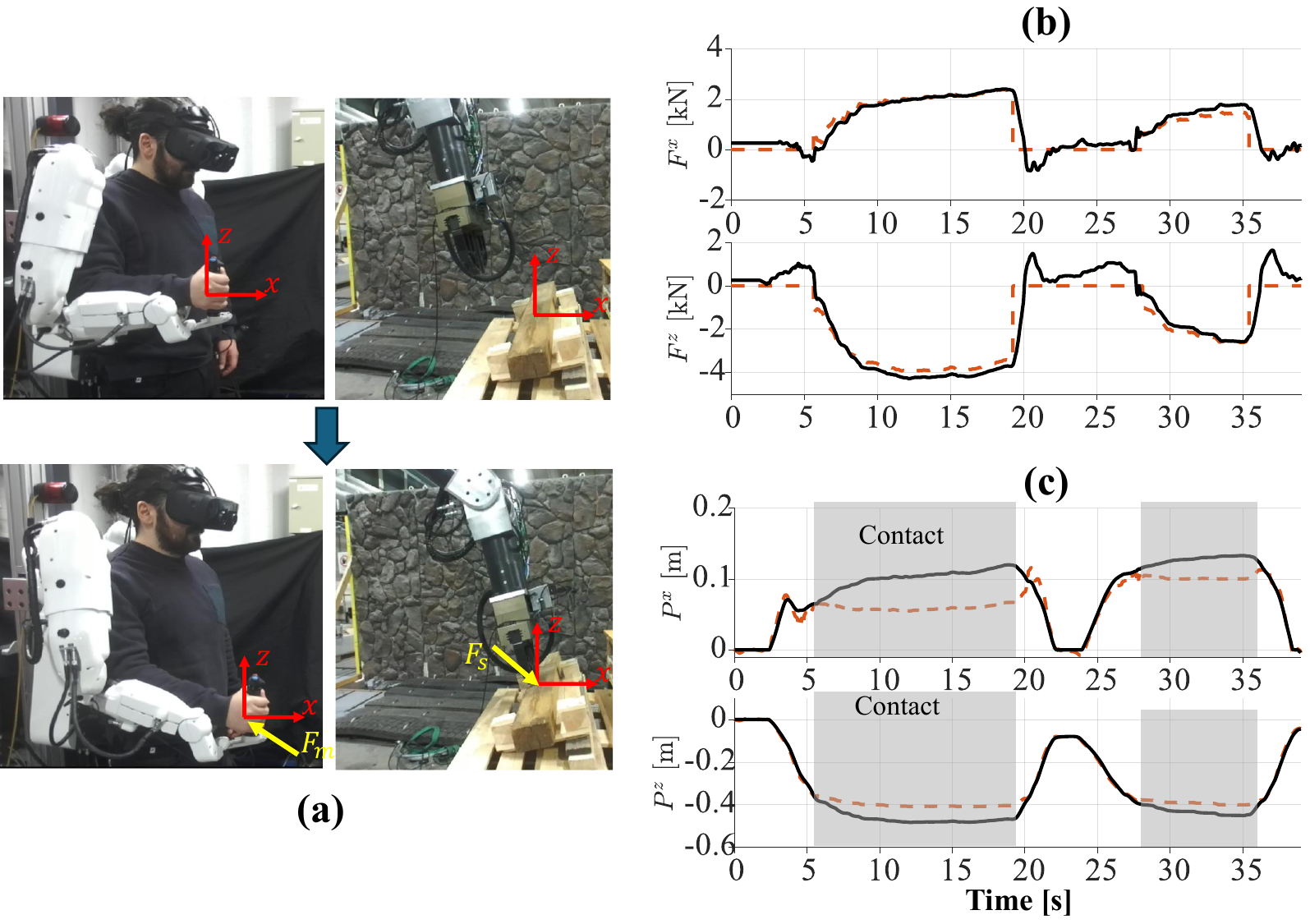}
\caption{{Multi-axis contact experiment with \mbox{\(\kappa_p = 2\)} and \mbox{\(\kappa_f = 200\)}. The red line represents the surrogate robot, and the black line indicates the master robot. (a) Experimental setup and task sequence; (b) force tracking in the x and z axes; and (c) position signals in the x and z axes.}}
\label{MA test}
\end{figure}

The results of the first set are presented in Fig.~\ref{contact_task}. As shown, the forces applied by the operator are effectively transmitted through the system and exerted on the remote environment with high fidelity. These results demonstrate that human-applied forces can be amplified up to 1000 times, which is a significant achievement. The second set of experiments were conducted in the presence of both fixed and time-varying communication delays, using a force scaling factor of \(\kappa_f = 800\). The results are shown in Fig.~\ref{contact_delay}: Fig.~\ref{contact_delay}(a) presents position and force tracking under a fixed delay, while Fig.~\ref{contact_delay}(b) shows the same under a time-varying delay. These delayed experiments were performed over extended durations compared to the non-delayed ones to further investigate system stability and transparency. Despite the presence of considerable communication delay, the proposed control scheme maintained a favorable trade-off between transparency and stability.

{The multi-axis contact experiment, shown in Fig.~\mbox{\ref{MA test}}(a), demonstrates the ability of the proposed system to handle complex force interactions in contact-rich environments. In this setup, the human operator maneuvers the surrogate robot to make contact with an inclined surface, allowing interaction along both the x- and z-axes. Upon contact, the proposed algorithm reflects the estimated environmental forces back to the operator in real time, enabling the user not only to observe but also to physically perceive the interaction through the force-reflective interface. As illustrated in Fig.~\mbox{\ref{MA test}}(b), the operator applied approximately 2~kN of force along the x-axis and 4~kN along the z-axis. The system effectively transmitted this intended action to the remote environment, demonstrating precise control and stable force execution. Fig.~\mbox{\ref{MA test}}(c) shows that the end-effector position was accurately tracked before and after contact. During contact, while the motion of the surrogate robot was constrained by the environment, the operator retained full control of the master device to apply intentional force along the constrained directions. This experiment highlights the controller’s robustness in rendering multi-directional contact forces and maintaining task continuity under physical constraints. It serves as clear evidence of the system’s capacity for immersive, high-fidelity bilateral teleoperation, delivering both visual and haptic feedback in challenging, real-world scenarios.}

\begin{rmk}
    It should be noted that both the human-applied forces and the contact forces at the remote site are estimated, which may introduce some inaccuracies, particularly on the master side. However, these estimation errors do not hinder the effectiveness of force reflection within the teleoperation system. As a result, the proposed force-sensor-less method offers a practical solution for real-world deployment, especially considering that conventional 6-DOF force/torque sensors are often fragile and susceptible to overloading.
\end{rmk}

A quantitative summary of the results is provided in Table~\ref{tab:contact_metrics}. For instance, with \(\kappa_f = 500\), the maximum and RMS force errors reached 737~N and 387~N, respectively. When the force scaling was increased to \(\kappa_f = 1000\), the maximum and RMS errors remained comparable at 848~N and 448~N, respectively, indicating robust tracking performance even under extreme amplification. As expected, the addition of delay degraded tracking performance to some extent. Nevertheless, the controller maintained stable execution and preserved a strong sense of transparency. Other metrics reported in Table~\ref{tab:contact_metrics}, such as position and orientation errors, further confirm the excellent performance of the controller in both delayed and non-delayed conditions. {Metrics in Table\mbox{~\ref{tab:contact_metrics}} further demonstrates the accuracy and stability of the algorithm in multi-axes experiment.}

\begin{table}[t]
\centering
\caption{Performance metrics in contact tasks}
\label{tab:contact_metrics}
\scriptsize
\begin{tabular}{p{0.22\columnwidth}cccc}
\toprule
\textbf{Experiment} & $\mathscr{N}_{p,\mathrm{rms}}$ [m] & $\mathscr{N}_{o,\mathrm{rms}}$ [deg] & $\mathscr{F}_{\mathrm{max}}$ [N] & $\mathscr{F}_{\mathrm{rms}}$ [N] \\
\midrule
$k_f = 500$ & {0.023} & 0.52 & 737.79 & 387.90 \\
$k_f = 800$ & {0.027} & 0.51 & 779.00 & 391.65 \\
$k_f = 1000$ & {0.028} & 0.45 & 848.60 & 448.25 \\
\midrule
\multicolumn{5}{l}{\textit{Experiments below are conducted with delay and \(\kappa_f = 800\):}} \\
\midrule
Fixed delay \(\mathcal{T}_{f} = 150~\mathrm{ms}\) & 0.10 & 0.75 & 1638.73 & 676.80 \\
Varying delay \(0 < \mathcal{T}_{v} < 150~\mathrm{ms}\) & 0.04 & 0.74 & 1171.71 & 588.88 \\
\midrule
\multicolumn{5}{l}{\textit{{Multi-axes contact experiment:}}} \\
\midrule
{\mbox{\(\kappa_f = 200\)}} & {0.0490} & {0.57} & {708.23} & {227.29} \\
\bottomrule
\end{tabular}
\normalsize
\end{table}

\subsection{User Study}
\label{User study section}
To evaluate the SoE and usability of the proposed teleoperation system, we conducted a user study using an adapted version of the questionnaire from~\cite{gonzalez2018avatar} for SoE assessment, alongside a custom-designed questionnaire to evaluate system usability. A total of 10 participants (9 male, 1 female) took part in the study, {where the participants were instructed to accomplish the RT experiment with \mbox{\(\kappa_p = 2\)} in two scenarios: one with VR-based visual feedback and the other with stereo-type camera information visualized on a monitor. Fig. \mbox{\ref{user}} illustrates the setup of experiments. The participant pool included individuals with varying anthropometric profiles, particularly in height, thereby allowing us to assess the ergonomic and functional adaptability of the teleoperation setup across different operator physiques and genders.} To intentionally challenge the system’s robustness under non-expert operation, the training phase was deliberately minimal. Participants were briefly introduced to the system, including expected haptic and visual feedback, and were then given only 2–3 minutes to familiarize themselves with the haptic exoskeleton through self-guided exploration. Such user studies are rarely reported in the literature, particularly for bilateral teleoperation systems involving real-world HHMs, which are typically characterized by nonlinearities, delays, and high inertial dynamics. To the best of our knowledge, very few, if any existing, studies have demonstrated comparable usability in such complex settings. Conducting a user study with non-expert participants on a large-scale system of this kind is especially unique and requires a highly reliable and safe experimental setup to ensure both user safety and meaningful evaluation. {At the conclusion of the experiment, participants completed a standardized 7-point Likert-scale questionnaire (1 = strongly disagree, 7 = strongly agree) designed to assess the SoA, SoSL, SoO, and the usability of the setup.} Although SoO was not a primary focus of this study, we included two items related to ownership to explore whether high SoA and SoSL might induce a partial SoO. The completed questionnaire is provided in Table~\ref{tab:questionnaire}. 

\begin{figure}[pos=h]
\centering
\includegraphics[width=.35\textwidth]{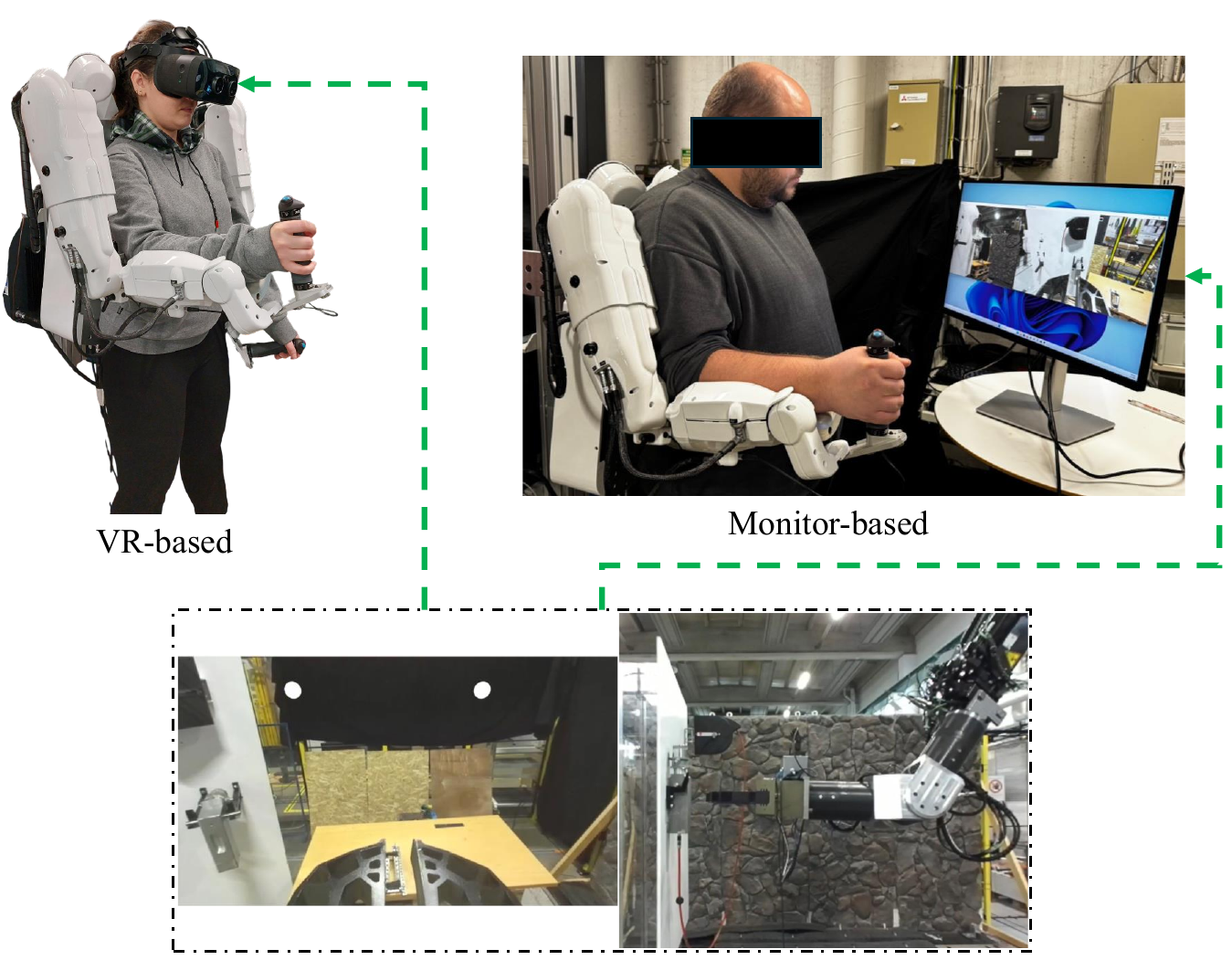}
\caption{{User study experiment with and without VR. The user is tasked to grasp the object on the white board.}} 
\label{user}
\end{figure}


\begin{table}[h!]
\scriptsize  
\centering
\caption{Questionnaire items and results for SoE and usability assessment}
\label{tab:questionnaire}
\begin{tabular}{p{0.1\columnwidth} p{0.38\columnwidth} p{0.16\columnwidth} p{0.16\columnwidth}}
\hline
\textbf{Q} & \textbf{Statement} & \textbf{VR} & \textbf{Monitor} \\
\hline
\multicolumn{4}{l}{\textbf{SOA}} \\
$Q_{A_1}$ & It felt like I could control the remote arm as if it was my own arm. & {$5.2 \pm 0.72$} & {$4.1 \pm 1.52$} \\ 
$Q_{A_2}$ & The movements of the remote arm were caused by my movements. & {$6.50 \pm 0.84$} & {$5.8 \pm 1.14$} \\ 
$Q_{A_3}$ & I felt as if the movements of the remote arm were influencing my own movements. & {$4.9 \pm 1.36$} & {$4.4 \pm 0.97$} \\
$Q_{A_4}$ & I felt as if the remote arm was moving by itself. & {$1.6 \pm 0.67$} & {$1.4 \pm 0.52$} \\[5pt]

\multicolumn{4}{l}{\textbf{SoSL}} \\
$Q_{L_1}$ & I felt as if I were located where the surrogate was. & {$5.4 \pm 0.80$} & {$3.1 \pm 1.37$} \\ 
$Q_{L_2}$ & It felt as if I was seeing the environment through the eyes of the surrogate. & {$6.0 \pm 0.95$} & {$3.6 \pm 1.71$} \\ 
$Q_{L_3}$ & I felt as if I was looking around the environment with my own head movements. & {$6.2 \pm 1.39$} & {$1.0$} \\[5pt]

\multicolumn{4}{l}{\textbf{SoO}} \\
$Q_{O_1}$ & I felt the remote body as if it were part of my own body. & {$4.9 \pm 1.14$} & {$3.7 \pm 1.57$} \\ 
$Q_{O_2}$ & I felt the remote body as an extension of my own body. & {$4.9 \pm 1.31$} & {$4.3 \pm 1.42$} \\[5pt]

\multicolumn{4}{l}{\textbf{Usability}} \\
$Q_{U_1}$ & I found the system very difficult to use. & {$2.00 \pm 0.47$} & {$2.3 \pm 1.34$} \\ 
$Q_{U_2}$ & I felt comfortable using the system. & {$5.70 \pm 0.89$} & {$4.8 \pm 1.14$} \\ 
$Q_{U_3}$ & I felt very confident using the system. & {$6.1 \pm 1.08$} & {$4.7 \pm 0.48$} \\
$Q_{U_4}$ & The task required a lot of mental effort. & {$2.70 \pm 1.41$} & {$2.8 \pm 1.03$} \\[5pt]
\hline
\end{tabular}
\end{table}

The results of the questionnaire in Table~\ref{tab:questionnaire} highlight the effectiveness of the proposed teleoperation setup in inducing strong SoA, self-location, and usability, despite the system's complexity and dissimilarity along with short familiarization period provided to the participants. Notably, participants reported a high level of control over the surrogate (\(Q_{A_1}\)) and a very low agreement with the statement that the robot moved autonomously (\(Q_{A_4}\)), indicating a robust SoA. Additionally, responses to \(Q_{A_1}\) and \(Q_{A_3}\) indicate that participants not only felt in control of the remote arm, but also experienced a tangible connection between their actions and the robot’s movements, likely facilitated by the distribution of scaled impedance of remote arm onto their own via the haptic exoskeleton. The high scores in the self-location category demonstrate that participants strongly felt as if they were spatially co-located with the surrogate. This is especially noteworthy given the asymmetry between the human operator’s body and the HHM, and it can be attributed to the use of egocentric viewpoint tracking and immersive VR headsets. Although inducing a sense of ownership was not a primary objective, the responses to \(Q_{O_1}\) and \(Q_{O_2}\) (~4.9 average) indicate that a moderate level of ownership was achieved. This is likely a result of the enhanced agency and self-location, both of which are known to contribute to the emergence of ownership in embodiment research. Interestingly, participants rated both the perception of the robot as a part of their body and as an extension of their body nearly equally. 

{In contrast to the VR condition, the monitor-based experiment demonstrated a notable reduction in participants' perceived embodiment. As shown in Table~\mbox{\ref{tab:questionnaire}}, the SoSL scores dropped significantly: \mbox{\(Q_{L_1}\)} decreased from \mbox{\(5.4 \pm 0.80\)} to \mbox{\(3.1 \pm 1.37\)}, and \mbox{\(Q_{L_2}\)} from \mbox{\(6.0 \pm 0.95\)} to \mbox{\(3.6 \pm 1.71\)}, indicating an approximate 50\% decline. Notably, one of the primary contributors to SoSL, \mbox{\(Q_{L_3} = 6.2 \pm 1.39\)} under VR, was entirely diminished under the monitor condition, as the monitor does not support head-motion-based egocentric view control. This reduction in SoSL appears to negatively influence SoA scores as well, likely due to the diminished immersion and reduced congruence between the user’s actions and the system’s visual response. It is also worth noting that the task completion time increased from \mbox{\(25.18 \pm 6.7\)} seconds in the VR condition to \mbox{\(45.48 \pm 12.19\)} seconds when using the monitor, indicating a substantial decline in overall task efficiency. This suggests that a lack of spatial embodiment may hinder the user's ability to fully interpret and internalize the robot’s motion as their own, resulting in a longer decision-making process and increased completion time.}

{To quantify the overall perceived SoE, the questionnaire responses were linearly normalized to a 0–100 scale. The resulting average normalized SoE score was $76.65\%$ for the VR condition and \mbox{$51.11\%$} for the monitor condition, indicating a relative improvement of \mbox{$50\%$} when using VR. This substantial enhancement is particularly noteworthy given the complexity and asymmetry of the master–surrogate robotic system employed. The usability results further support that the improved SoE was achieved without adding complexity to the system. Usability scores indicate that users felt more comfortable and experienced lower mental workload when using the VR interface compared to the monitor-based setup for the same task. These scores underscore the intuitiveness and accessibility of the system, even for users with no prior experience without limitation to gender, further validating the ergonomic and cognitive effectiveness of proposed control and interface design.}

\section{Conclusion}\label{conclusion}

{This study introduced an immersive force-sensorless bilateral teleoperation framework with force reflection, designed for highly asymmetric, beyond-human-scale manipulators. To tackle the core challenge of operator embodiment, we focused on enhancing the SoA and SoSL through both control and interface design. A robust controller ensured high transparency under severe model uncertainties, actuator nonlinearities, and arbitrary time delays, with theoretical guarantees of semi-global uniform ultimate boundedness. Immersion was reinforced using a wearable haptic exoskeleton with distributed impedance rendering and a VR headset enabling egocentric head tracking. Extensive experiments validated the system’s performance across unprecedented scaling conditions (\mbox{\(1{:}13\)} motion, \mbox{\(1{:}1000\)} force), as well as under fixed and time-varying delays. The newly introduced CPT task showcased precise 6-DoF manipulation in complex environments, while multi-axis contact experiments confirmed accurate force reflection and stability in constraint-rich interactions. A detailed user study, including a VR ablation comparison, revealed a 50\mbox{\%} improvement in normalized embodiment scores with VR, reaching an average of 76.65\mbox{\%}. Notably, users reported high comfort and confidence, highlighting the system's intuitiveness despite the scale and dynamics of the heavy-duty surrogate. Together, these results establish a compelling foundation for scalable, immersive, and embodied remote manipulation.}

{Future developments will include a user study in contact-rich scenarios with trained participants to quantify the specific impact of force reflection on SoE.} Another promising direction involves mapping contact points on the surrogate to anatomically equivalent regions on the human body to enhance embodiment. We also aim to develop shared control strategies to improve task efficiency, optimize the control and placement of the pan-tilt camera system for enhanced spatial congruence, and explore learning-from-demonstration techniques to support skill transfer and partial autonomy in heavy-duty systems.

\end{defka}

\section*{Appendix}
\appendix
\renewcommand{\theequation}{\thesection.\arabic{equation}}

\section{Boundedness of \(p_{T_7}\)}
\setcounter{equation}{0}
\label{p_T_7}

In the context of VDC, the augmented spatial velocity vector is defined as
\begin{equation}\label{AugV}
    \mathscr{V} = \mathscr{J}\Dot{\mathrm{q}},
\end{equation}
where \(\Dot{\mathrm{q}} \in \Re^7\) is the vector of joint velocities, and \(\mathscr{J} \in \Re^{49\times7}\) is the augmented Jacobian matrix that maps the joint velocity vector to the augmented spatial velocity vector. \(\mathscr{V} \in \Re^{49}\) contains the spatial velocities of all frames in the decomposed system (for more details, refer to~\cite[Section 3.3.5]{zhu2010virtual}). 

Similarly, the following expression maps the augmented spatial force vector of the decomposed system to the joint torque vector:
\begin{equation}\label{AugF}
    \tau = \mathscr{J}^T \mathcal{F},
\end{equation}
where \(\mathcal{F} \in \Re^{49}\) and \(\tau \in \Re^7\) denote the joint torque vector. Exploiting (\ref{AugV}) and (\ref{AugF}), the following expression for the human arm dynamics can be reconstructed from (\ref{HR body})–(\ref{tau}):
\begin{equation}
\begin{split}
    \tau_m &= \mathscr{J}^T \mathscr{M}_h\mathscr{J}\, \Ddot{\mathrm{q}} +\mathscr{J}^T \mathscr{M}_h\Dot{\mathscr{J}}\, \Dot{\mathrm{q}} + \mathscr{J}^T \mathscr{C}_h + \tau_h,
\end{split}
\label{Fh}
\end{equation}
where \(\mathscr{M}_h \in \Re^{49\times49}\) and \(\mathscr{C}_h \in \Re^{49}\) are the augmented mass and damping matrices of the human arm, \(\tau_h \in \Re^{7}\) is the exogenous torque, and \(\tau_m \in \Re^{7}\) represents the human–robot interaction torque. In the same sense, we can write:
\begin{equation}\label{t_mr}
    \tau_{mr} = \mathscr{J}^T \hat{\mathscr{M}}_h\mathscr{J}\, \Ddot{\mathrm{q}}_r +\mathscr{J}^T \hat{\mathscr{M}}_h\Dot{\mathscr{J}}\, \Dot{\mathrm{q}} + \mathscr{J}^T \hat{\mathscr{C}}_h + \hat{\tau}_h.
\end{equation}
The linear-in-parameter form for (\ref{t_mr}) can be written as:
\begin{equation}
    \mathscr{Y}_h \hat{\Phi}_h = \mathscr{J}^T \hat{\mathscr{M}}_h\mathscr{J}\, \Ddot{\mathrm{q}}_r +\mathscr{J}^T \hat{\mathscr{M}}_h\Dot{\mathscr{J}}\, \Dot{\mathrm{q}} + \mathscr{J}^T \hat{\mathscr{C}}_h,
\end{equation}
where \(\mathscr{Y}_h \in \Re^{7\times49}\) is the augmented regression matrix and \(\hat{\Phi}_h \in \Re^{49}\) is the estimated augmented inertial parameter vector of the human arm. Note that Definition~\ref{121 map} can be used for \(\hat{\Phi}_h\).

By adding and subtracting \(\mathscr{J}^T {\mathscr{M}}_h\mathscr{J}\, \Ddot{\mathrm{q}}_r\) to the right-hand side of (\ref{Fh}) and substituting into \((\tau_{mr} - \tau_{m})\), we obtain:
\begin{equation}
\begin{split}
    \tau_{mr} - \tau_{m} &= \mathscr{Y}_h (\hat{\Phi}_h - \Phi_h) + \mathscr{J}^T {\mathscr{M}}_h\mathscr{J} (\Ddot{q}_r - \Ddot{\mathrm{q}}) + \hat{\tau}_h - \tau_h.
\end{split}
\label{dif tm}
\end{equation}
Assuming that the human arm and the robot have the same velocity at the end-effector, the VPF at frame \(\{T_7\}\) of the master robot can be written as:
\begin{equation}
\begin{split}\label{pt7}
    p_{T_7} &= (\,^{T_7}V_r - \,^{T_7}V)^T (\,^{T_7}F_r - \,^{T_7}F) \\
    &= (\Dot{X}_{hr} - \Dot{X}_{h})^T (F_{mr} - F_m),
\end{split}
\end{equation}
with
\begin{equation}\label{Dxhh}
    \Dot{X}_{h} = V_{m} = \mathcal{J}_m \Dot{\mathrm{q}}, \quad \Dot{X}_{hr} = V_{mr} = \mathcal{J}_m \Dot{\mathrm{q}}_r,
\end{equation}
\begin{equation}\label{Fmm}
    \tau_{m} = \mathcal{J}_m^T F_m, \quad \tau_{mr} = \mathcal{J}_m^T F_{mr}.
\end{equation}
During physical human–robot interaction, \(p_{T_7}\) represents the power injected or absorbed by the human operator. Additionally, during the interaction phase, as the human grasps the handle of the exoskeleton, there is a continuous flow of power between the robot and the human arm, indicating that \(||p_{T_7}|| \geq \varrho_m\), where \(\varrho_m > 0\) is a small constant.

By substituting (\ref{Dxhh}) and (\ref{Fmm}) into (\ref{pt7}), we obtain:
\begin{equation}
\begin{split}\label{pt7_1}
    p_{T_7} = (\Dot{\mathrm{q}}_r - \Dot{\mathrm{q}})^T (\tau_{mr} - \tau_m).
\end{split}
\end{equation}
Substituting (\ref{dif tm}) into (\ref{pt7_1}) gives:
\begin{equation}
\begin{split}\label{pt7_2}
    p_{T_7} &= (\Dot{\mathrm{q}}_r - \Dot{\mathrm{q}})^T \mathscr{Y}_h (\hat{\Phi}_h - \Phi_h) \\
    &+ (\Dot{\mathrm{q}}_r - \Dot{\mathrm{q}})^T \mathscr{J}^T {\mathscr{M}}_h \mathscr{J} (\Ddot{\mathrm{q}}_r - \Ddot{\mathrm{q}}) \\
    &+ (\Dot{\mathrm{q}}_r - \Dot{\mathrm{q}})^T (\hat{\tau}_h - \tau_h).
\end{split}
\end{equation}
Taking the integral of \(p_{T_7}\) yields:
\begin{equation}
\begin{split}
    \lim_{t \to \infty} \int_0^{t} p_{T_7} \,dt &= \lim_{t \to \infty} \int_0^{t} (\Dot{\mathrm{q}}_r - \Dot{\mathrm{q}})^T \mathscr{Y}_h (\hat{\Phi}_h - \Phi_h) \,dt \\
    &+ \lim_{t \to \infty} \int_0^{t} (\Dot{\mathrm{q}}_r - \Dot{\mathrm{q}})^T \mathscr{J}^T {\mathscr{M}}_h \mathscr{J} (\Ddot{\mathrm{q}}_r - \Ddot{\mathrm{q}}) \,dt \\
    &+ \lim_{t \to \infty} \int_0^{t} (\Dot{\mathrm{q}}_r - \Dot{\mathrm{q}})^T (\hat{\tau}_h - \tau_h) \,dt,
\end{split}
\end{equation}
which leads to:
\begin{equation}
    \lim_{t \to \infty} \int_0^{t} p_{T_7} \,dt \geq -\varrho_{0m},
\end{equation}
with
\begin{equation}
\begin{split}
    \varrho_{0m} &= \dfrac{\gamma}{2} \Tilde{\Phi}_h(0)^T \Gamma(\hat{\Phi}_h(0)) \Tilde{\Phi}_h(0) \\
    &+ \dfrac{1}{2} (\Dot{\mathrm{q}}_r(0) - \Dot{\mathrm{q}}(0))^T \mathscr{J}^T \mathscr{M}_h \mathscr{J} (\Dot{\mathrm{q}}_r(0) - \Dot{\mathrm{q}}(0)),
\end{split}
\end{equation}
where \(\Gamma(\hat{\Phi}_h(0))\) denotes the constant pullback of the Riemannian metric~\cite{hejrati2022decentralized}.

\section{Proof of Lemma \ref{pT} }
\setcounter{equation}{0}
\label{p_T}
        By employing Definition \ref{Def pA}, (\ref{f_e}) and (\ref{f_ed}) along with \(V_s = V_T\), the following holds for \(p_{T}\),
        \begin{equation}
        \begin{split}
            p_T &= (^TV_r-\,^TV)^T(^TF_r-\,^TF)\\
            & = (^TV_r-\,^TV)^T\,N_c\,(f_{ed}-f_e)\\
            & = (^TV_r-\,^TV)^T\,N_c\,M_e (\Dot{V}_{sr}-\Dot{V}_{s})\\
            & = (V_{sr}-\,V_s)^T\,M_e (\Dot{V}_{sr}-\Dot{V}_{s}),
        \end{split}
        \end{equation}
        which leads to,
        \begin{equation}
            \begin{split}
            \int_0^\infty {p_Tdt} & =  \int_0^\infty {(V_{sr}-\,V_s)^T\,M_e (\Dot{V}_{sr}-\Dot{V}_{s})dt} \\
            & \geq -0.5 (V_{sr}(0)-\,V_s(0))^T\,M_e ({V}_{sr}(0)-{V}_{s}(0))\\
            & \geq - \varrho_{0s}.
        \end{split}
        \end{equation}

\bibliographystyle{elsarticle-num}
\bibliography{mybib}

\end{document}